\documentclass{article}

\PassOptionsToPackage{numbers, compress}{natbib}
\usepackage[preprint]{neurips_2026}

\usepackage[utf8]{inputenc} 
\usepackage[T1]{fontenc}    
\usepackage{hyperref}       
\usepackage{url}            
\usepackage{booktabs}       
\usepackage{amsfonts}       
\usepackage{nicefrac}       
\usepackage{microtype}      
\usepackage{xcolor}         

\usepackage{multirow}
\usepackage{amsmath}
\usepackage{amssymb}
\usepackage{graphicx}
\usepackage{colortbl}
\usepackage{wrapfig}
\usepackage{algorithm}
\usepackage{algpseudocodex}

\definecolor{myframepink}{RGB}{72,138,176}

\hypersetup{
    colorlinks=true,       
    allcolors=myframepink  
}

\title{Gradient-Informed Temporal Sampling Improves Rollout Accuracy in PDE Surrogate Training}

\author{
Wenshuo Wang$^{1}$, Fan Zhang$^{2,3}$\thanks{Corresponding author.} \\
$^{1}$ School of Future Technology, South China University of Technology, China \\
$^{2}$ State Key Laboratory of Ocean Sensing \& Ocean College, Zhejiang University, China \\
$^{3}$ Kavli Institute for Astrophysics and Space Research, Massachusetts Institute of Technology, USA \\
\texttt{202364870251@mail.scut.edu.cn}, \texttt{f.zhang@zju.edu.cn}
}

\begin{document}

\maketitle

\begin{abstract}
Researchers train neural simulators on uniformly sampled numerical simulation data. But under the same budget, does systematically sampled data provide the most effective information? A fundamental yet unformalized problem is how to sample training data for neural simulators so as to maximize rollout accuracy. Existing data sampling methods either tend to collapse into locally high-information-density regions, or preserve diversity but remain insufficiently model-specific, often leading to performance that is no better than uniform sampling. To address this, we propose a data sampling method tailored to neural simulators, Gradient-Informed Temporal Sampling (GITS). GITS jointly optimizes pilot-model local gradients and set-level temporal coverage, thereby effectively balancing model specificity and dynamical information. Compared with multiple sampling baselines, the data selected by GITS achieves lower rollout error across multiple PDE systems, model backbones and sample ratios. Furthermore, ablation studies demonstrate the necessity and complementarity of the two optimization objectives in GITS. In addition, we analyze the successful sampling patterns of GITS as well as the typical PDE systems and model backbones on which GITS fails.
\end{abstract}

\section{Introduction}

Neural simulators are trained on continuous-time physical-field trajectories generated by numerical solvers, and then serve as surrogates for PDE evolution~\cite{takamoto2022pdebench,li2020fourier}.
In existing practice, researchers usually train on temporally contiguous data sampled at equal time intervals~\cite{ohana2024well,mccabe2024multiple}.
However, a persistent yet never systematically studied problem is that such systematic sampling does not always provide the most informative or the most dynamics-relevant training pairs.
This raises a basic question: given an existing pool of numerical simulation data, how should one sample from it so that a neural simulator trained under the same budget achieves the best rollout performance? In this paper, we study this question in the offline shared temporal start-index setting, where a common subset of admissible start indices is selected once from a fixed, fully generated trajectory dataset and then reused across all training trajectories.
As model capacity, parameter scale, and training techniques continue to improve, the quality and informativeness of the training data itself become increasingly important to the final performance of neural simulators~\cite{herde2024poseidon,chen2024data}.

One of the most widely used families of data valuation methods is pilot-model, gradient-driven subset selection~\cite{killamsetty2021glister,killamsetty2021grad,guo2022deepcore}.
Such methods can effectively estimate the information gain of individual samples for a model, but they do not explicitly account for information redundancy among samples~\cite{paul2021deep,ash2019deep}.
As a result, when applied to spatiotemporal data such as PDE trajectories, they tend to select densely clustered neighboring time steps~\cite{kim2025flexible,jantre2025data}.
On the other hand, common coverage-based subset selection methods face the opposite limitation: they balance diversity at the set level, but are largely model-agnostic, and therefore cannot identify which temporal windows are intrinsically the most useful for downstream neural surrogate training~\cite{kothawade2022prism,sener2017active,zheng2022coverage,maharana2023d2}.

To address this gap, we propose \textbf{Gradient-Informed Temporal Sampling (GITS)}, a sampling method tailored to offline PDE surrogate training on pre-generated simulation data. GITS jointly optimizes two complementary objectives: (1) a low-cost pilot-model short-horizon gradient-norm score that serves as a practical local proxy for candidate usefulness, and (2) a set-level temporal coverage objective that encourages the selected subset to cover the temporal axis rather than collapse onto highly redundant neighboring regions. Together, they overcome the respective weaknesses of pure score-based ranking and pure coverage-based selection.

Using three representative PDE systems from PDEBench, we show that data selected by GITS leads to lower rollout error than a broad set of temporal selection baselines across several mainstream neural simulator architectures and sample ratios~\cite{takamoto2022pdebench}. Furthermore, our ablation studies demonstrate necessity of the two components of GITS. Finally, we conduct detailed analyses of the sampling patterns through which GITS reduces error relative to the baselines, as well as the failure boundaries in the minority of cases where it underperforms.
In summary, the contributions of this paper are:
\begin{itemize}
    \item We formalize and explicitly study the overlooked problem of offline shared temporal-window selection for autoregressive neural PDE surrogate training under a fixed budget.
    \item We propose GITS, which combines a pilot-model short-horizon gradient proxy with set-level temporal coverage regularization, and show that it improves over strong temporal selection baselines on representative PDEBench tasks.
    \item We clarify when and why GITS helps or fails, showing that its behavior is governed by score--utility alignment and by whether temporal coverage is beneficial or counterproductive.
\end{itemize}

\section{Related Work}

\subsection{Budgeted Temporal Window Selection for Neural PDE Surrogate Training}

We formalize \emph{budgeted temporal window selection for neural PDE surrogate training} as follows.
Consider a collection of $N$ generated PDE trajectories
$\{x_t^{(n)}\}_{t=0}^{T_c-1}$, where $x_t^{(n)}$ denotes the system state of trajectory $n$ at coarse time step $t$, and $T_c$ is the number of available coarse time steps after any preprocessing or temporal downsampling.
Let $L$ denote the history length used by an autoregressive surrogate to predict the next state.
Then, each admissible temporal start index $k$ induces a training window consisting of an input history $x_{k-L+1:k}^{(n)}$ and a one-step target $x_{k+1}^{(n)}$, where $x_{a:b}^{(n)}$ denotes the contiguous subsequence from time step $a$ to $b$.
Accordingly, the candidate start-index set is
\[
\mathcal{C}=\{k\in\mathbb{Z}\mid L\le k\le T_c-2\}.
\]
Under a budget $K$, the task is to select a subset $\mathcal{S}\subseteq\mathcal{C}$ with $|\mathcal{S}|=K$, which induces the training set
\[
\mathcal{D}(\mathcal{S})
=
\left\{
\bigl(x_{k-L+1:k}^{(n)},\,x_{k+1}^{(n)}\bigr)
\;\middle|\;
n=1,\dots,N,\; k\in\mathcal{S}
\right\},
\]
such that a surrogate trained on $\mathcal{D}(\mathcal{S})$ achieves the best possible downstream rollout accuracy under the same budget. Because the same start-index subset $\mathcal{S}$ is reused across all trajectories, this is a \emph{shared} temporal start-index formulation rather than a trajectory-specific selector.
Unlike active data acquisition, this problem assumes that the full trajectories already exist: the question is not which simulations or timestamps to query, but which temporal windows to retain for training.

This problem is persistent because mainstream autoregressive neural PDE surrogate pipelines are typically trained on uniformly discretized trajectories, which treats all admissible temporal windows as equally useful for training.
Recent data-efficiency efforts have begun to move beyond brute-force full-data usage, but mostly at other levels: active-learning frameworks for neural PDE solvers focus on selecting informative trajectories or parametric PDE instances in a solver-in-the-loop setting \cite{musekamp2025active}; selective time-step acquisition chooses which states to generate during simulation \cite{pmlr-v267-kim25m}; continuous-time or irregular-time models aim to learn from non-uniform temporal grids \cite{hou2026cfo}; and coreset-style approaches select informative inputs or samples for simulation and labeling \cite{satheesh2025picore}.
To the best of our knowledge, prior work has not isolated the offline problem defined above as a standalone task.

\subsection{Gradient-Based Data Valuation and Coverage-Based Subset Selection}

Among the broad literature on training subset selection, the two technical families most relevant to our setting are gradient-based data valuation and coverage-based subset selection~\cite{guo2022deepcore,zheng2022coverage}.

Gradient-based data valuation methods can be broadly divided into local score-based ranking and gradient-driven subset selection~\cite{guo2022deepcore}.
GraNd uses gradient-derived local scores to identify training examples that are important early in optimization \cite{paul2021datadiet}.
GRAD-MATCH selects subsets by matching the gradient of the selected subset to that of the full training or validation data \cite{killamsetty2021gradmatch}.
GLISTER instead formulates data subset selection as a validation-oriented objective, providing a model-aware criterion for selecting effective subsets \cite{killamsetty2021glister}.
These methods are attractive because they offer computable proxies for data usefulness that depend on the model's own optimization signal.
However, for the budgeted temporal window selection task defined in the previous subsection, they do not explicitly address the strong redundancy that may arise among nearby temporal windows, and therefore cannot by themselves guarantee a temporally well-distributed selected subset.

Coverage-based subset selection methods can likewise be divided into classical representative coverage objectives and more guided subset selection objectives~\cite{wei2015submodularity,kothawade2022prism}.
Classical submodular formulations such as facility-location evaluate how well a selected subset covers the full ground set and admit efficient greedy optimization with approximation guarantees \cite{wei2015submodularity}.
More recent guided formulations, such as PRISM, further extend this perspective by parameterizing submodular information measures to balance coverage with additional selection desiderata \cite{kothawade2022prism}.
These methods are attractive because they provide principled set-level control over representativeness and diversity.
However, when used alone for the task defined above, they remain largely model-agnostic: they can encourage temporally spread selections, but they do not indicate which temporal windows are intrinsically the most useful for downstream neural surrogate training.

Motivated by the complementarity of these two technical families, GITS combines gradient-based data valuation with coverage-based subset selection and optimizes them jointly.

\section{Gradient-Informed Temporal Sampling}

In this section, we first outline the workflow of GITS in Section~\ref{sec:flow}, then detail its components in Sections~\ref{sec:com1} and~\ref{sec:com2}, with the optimization objective in Section~\ref{sec:com3}.

\subsection{Overview of Gradient-Informed Temporal Sampling}\label{sec:flow}

\begin{algorithm}[t]
\caption{Gradient-Informed Temporal Sampling (GITS)}
\label{alg:gits}
\begin{algorithmic}[1]
\Require Fully generated PDE trajectories $\{x_t^{(n)}\}_{t=0}^{T_c-1}$ for $n=1,\dots,N$, surrogate model class $f_\theta$, history length $L$, budget $K$
\Ensure Selected start-index set $\mathcal{S}$ with $|\mathcal{S}|=K$
\State Construct the candidate start-index set
\[
\mathcal{C}\gets \{k\in\mathbb{Z}\mid L\le k\le T_c-2\}
\]
\State Compute pointwise informativeness scores $\{s_k\}_{k\in\mathcal{C}}$
\State Construct the set-level temporal coverage terms
\State Greedily maximize the GITS objective under $|\mathcal{S}|=K$
\State \Return $\mathcal{S}$
\end{algorithmic}
\end{algorithm}

Algorithm~\ref{alg:gits} summarizes the workflow of GITS. Line~1 defines the admissible temporal start indices on the coarse trajectory grid. Line~2 assigns each candidate a model-aware pointwise score using a lightweight pilot model and a short-horizon rollout gradient. This provides a practical local proxy for usefulness, but by itself may favor many neighboring start indices with highly redundant dynamics. Line~3 therefore adds two set-level temporal coverage terms: a global term that spreads selections over the candidate axis and a sliding-window term that reduces local temporal gaps. Line~4 then greedily maximizes the resulting joint objective under the budget constraint.

\subsection{Pointwise informativeness estimation}\label{sec:com1}

To obtain low-cost model-aware candidate scores, GITS first trains a lightweight pilot model. Let $\mathcal{P}$ denote the pilot start pool. In GITS, we set $\mathcal{P}=\mathcal{C}$, i.e., the pilot is trained on the full candidate pool for $E_p$ epochs, and denote the resulting pilot parameters by $\theta_p$.

Let $H$ denote the pilot short-rollout horizon. For each candidate start index $k\in\mathcal{C}$, we define the effective rollout horizon
\[
H_k := \min(H,\, T_c-1-k),
\]
so that candidates near the end of the trajectory remain well-defined. For trajectory $n$, we initialize the rollout with the ground-truth history window
\[
\hat{x}_{k-L+1:k}^{(n)} = x_{k-L+1:k}^{(n)},
\]
where $L$ is the autoregressive history length. Starting from this history, we recursively predict for $h=1,\dots,H_k$,
\[
\hat{x}_{k+h}^{(n)} = f_\theta\!\left(\hat{x}_{k+h-L:k+h-1}^{(n)}\right).
\]
The candidate-specific short-rollout loss is
\[
\ell_k(\theta)
=
\frac{1}{H_k}\sum_{h=1}^{H_k}
\mathbb{E}_{n}\!\left[
\operatorname{NRMSE}\!\left(\hat{x}_{k+h}^{(n)},x_{k+h}^{(n)}\right)^2
\right],
\]
where $\operatorname{NRMSE}(\cdot,\cdot)$ denotes normalized root-mean-square error, and $\mathbb{E}_{n}[\cdot]$ denotes averaging over trajectories, or its mini-batch approximation in implementation.

At the pilot model, we compute
\[
g_k := \nabla_\theta \ell_k(\theta)\big|_{\theta=\theta_p},
\qquad
s_k := \|g_k\|_2,
\]
and use $s_k$ as the pointwise informativeness score of candidate $k$. Intuitively, $s_k$ measures how strongly training on start index $k$ would update the model around the pilot parameters, and thus serves as a practical local proxy for candidate usefulness.

\subsection{Set-level temporal coverage regularization}\label{sec:com2}

High pointwise scores alone do not prevent the selected subset from collapsing onto densely clustered neighboring time steps. GITS therefore adds two set-level temporal coverage terms on subsets $\mathcal{S}\subseteq\mathcal{C}$.

We first define a global temporal coverage term. For any $i,j\in\mathcal{C}$, let
\[
S_{ij}
=
\exp\!\left(-\frac{|i-j|}{\tau}\right),
\]
where $\tau>0$ is the global temporal decay scale. The corresponding global coverage is
\[
F_{\mathrm{cov}}(\mathcal{S})
=
\sum_{i\in\mathcal{C}} \max_{j\in\mathcal{S}} S_{ij}.
\]
This term encourages every admissible temporal location to be close to at least one selected start index.

To further reduce local temporal gaps, we define a sliding-window coverage term. Let
\[
\mathcal{I}=\{[a_m,b_m]\}_{m=1}^{M}
\]
be temporal windows generated on the candidate axis with window size $W$ and stride $S_w$, where $M$ is the number of windows. For window $m$ and candidate start index $j$, define the distance from $j$ to the interval $[a_m,b_m]$ by
\[
d(m,j)=
\begin{cases}
0, & j\in[a_m,b_m],\\
a_m-j, & j<a_m,\\
j-b_m, & j>b_m,
\end{cases}
\]
and define the window-level similarity
\[
R_{mj}
=
\exp\!\left(-\frac{d(m,j)}{\tau_w}\right),
\]
where $\tau_w>0$ is the local decay scale. The corresponding sliding-window coverage is
\[
F_{\mathrm{win}}(\mathcal{S})
=
\sum_{m=1}^{M} \max_{j\in\mathcal{S}} R_{mj}.
\]
The global term promotes temporal spread over the full candidate axis, while the sliding-window term improves local coverage regularity.

\subsection{Joint objective and greedy optimization}\label{sec:com3}

Given the pointwise scores $\{s_k\}_{k\in\mathcal{C}}$ and the two set-level coverage terms, GITS solves
\[
\max_{\mathcal{S}\subseteq\mathcal{C}} F(\mathcal{S})
\qquad
\text{s.t.}
\qquad
|\mathcal{S}|=K,
\]
where $K$ is the sampling budget, determined in experiments by the chosen sampling ratio. The objective is
\[
F(\mathcal{S})
=
\sum_{k\in\mathcal{S}} s_k
+
\lambda_{\mathrm{cov}}\,F_{\mathrm{cov}}(\mathcal{S})
+
c_{\mathrm{win}}\,F_{\mathrm{win}}(\mathcal{S}),
\]
where $\lambda_{\mathrm{cov}}\ge 0$ and $c_{\mathrm{win}}\ge 0$ are the weights of the global and sliding-window coverage terms, respectively. The first term favors individually informative start indices, while the latter two terms discourage redundant temporal concentration.

The objective admits an efficient greedy solution. The score term $\sum_{k\in\mathcal{S}} s_k$ is modular, and both $F_{\mathrm{cov}}(\mathcal{S})$ and $F_{\mathrm{win}}(\mathcal{S})$ are monotone submodular facility-location terms. Therefore, $F(\mathcal{S})$ is also monotone submodular, and the standard greedy algorithm yields a $(1-1/e)$-approximation under the cardinality constraint.

To compute greedy gains efficiently, we maintain the current coverage states
\[
m_i := \max_{j\in\mathcal{S}} S_{ij},
\qquad i\in\mathcal{C},
\]
and
\[
u_m := \max_{j\in\mathcal{S}} R_{mj},
\qquad m=1,\dots,M,
\]
with $m_i=0$ and $u_m=0$ when $\mathcal{S}=\varnothing$. For any unselected candidate $k\in\mathcal{C}\setminus\mathcal{S}$, its marginal gain is
\[
\Delta(k\mid\mathcal{S})
=
s_k
+
\lambda_{\mathrm{cov}}
\sum_{i\in\mathcal{C}}
\bigl(\max(m_i,S_{ik})-m_i\bigr)
+
c_{\mathrm{win}}
\sum_{m=1}^{M}
\bigl(\max(u_m,R_{mk})-u_m\bigr).
\]
GITS repeatedly selects
\[
k^\star = \arg\max_{k\in\mathcal{C}\setminus\mathcal{S}} \Delta(k\mid\mathcal{S}),
\]
updates $\mathcal{S}\leftarrow\mathcal{S}\cup\{k^\star\}$ together with $\{m_i\}$ and $\{u_m\}$, and stops when $|\mathcal{S}|=K$.

\section{Empirical Evaluation}

Our experiments are divided into three parts, designed to verify that:
\begin{itemize}
    \item \textbf{Effectiveness}: GITS improves rollout accuracy for PDE surrogate training;
    \item \textbf{Necessity}: the individual components of GITS are necessary; and
    \item \textbf{Boundary Conditions}: the success and failure regimes of GITS.
\end{itemize}

\subsection{Experimental setup}\label{sec:exp_setup}

\subsubsection{Datasets}\label{sec:datasets}

We evaluate on three PDEBench forward tasks chosen to span distinct physical regimes and boundary conditions rather than multiple variants of the same PDE family~\cite{takamoto2022pdebench}. Specifically, \texttt{diff-sorp} is a 1D scalar diffusion-sorption problem with Cauchy boundary conditions, \texttt{diff-react} is a 2D two-field reaction-diffusion system with Neumann boundaries, and \texttt{rdb} is the radial-dam-break instance of PDEBench's 2D shallow-water benchmark, representing hyperbolic free-surface dynamics with shock-capable wave propagation. 

According to PDEBench, these three tasks comprise 10{,}000, 1{,}000, and 1{,}000 trajectories, respectively~\cite{takamoto2022pdebench}. Their benchmark resolutions are 1024 for \texttt{diff-sorp} and $128\times128$ for both 2D tasks; \texttt{diff-react} and \texttt{rdb} each contain 100 temporal evolution steps (equivalently 101 stored snapshots when counting the initial frame), while \texttt{diff-sorp} has 100 temporal steps. \texttt{diff-react} contains two channels $(u,v)$, whereas \texttt{diff-sorp} and \texttt{rdb} are single-channel scalar-field prediction tasks.

\subsubsection{Baseline selection and implementation details}\label{sec:baseline}

At the backbone level, we evaluate four standard PDE surrogate backbones: U-Net, Fourier Neural Operator (FNO), ConvLSTM, and Transformer. All models are trained as one-step autoregressive predictors with history length \(L=4\). For U-Net, FNO, and Transformer, the \(L\) input frames are concatenated along the channel dimension; for ConvLSTM, the same history is fed as a sequence. Detailed architectural hyperparameters are provided in Appendix~\ref{app:extra_baselines}.

At the sampling level, we compare GITS against six baselines spanning standard uniform sampling, pointwise pilot-based scoring, set-level temporal coverage, and stronger generic subset selection. The \textit{uniform} baseline is the standard evenly spaced temporal sampler. The \textit{loss-only} baseline trains a lightweight pilot model of the same backbone, scores each candidate start by its short-rollout loss under the pilot, and keeps the top-\(K\) starts. The \textit{coverage-only} baseline removes the local utility term from the GITS objective and optimizes only the temporal coverage terms. \textit{GradMatch} is adapted from the original method to our shared temporal start-index setting~\citep{killamsetty2021gradmatch}: each candidate start is treated as an item, represented by its pilot-model short-rollout gradient aggregated over training trajectories, and selection approximately matches the full-candidate gradient using greedy gradient matching.

We also include task-adapted GLISTER-style and PRISM-style baselines~\cite{killamsetty2021glister,kothawade2022prism}. The \textit{GLISTER-style} baseline follows GLISTER's validation-guided approximation, adapted from i.i.d.\ example selection to shared temporal-start selection for autoregressive PDE rollout training. The \textit{PRISM-style} baseline instantiates a guided-submodular targeted-subset selector in the same setting, using validation-defined query/private sets and rollout-gradient CountSketch embeddings. All pilot-based samplers (\textit{loss-only}, \textit{GradMatch}, \textit{GLISTER-style}, \textit{PRISM-style}, and GITS) use the same pilot protocol as GITS, with the pilot start pool set to the full candidate set on the training trajectories. Detailed backbone settings and exact hyperparameter values for the pilot stage, the GITS objective, and the GLISTER-style and PRISM-style baselines are deferred to Appendix~\ref{app:gits_hparams} and Appendix~\ref{app:extra_baselines}.

All experiments were conducted on a server equipped with four NVIDIA A100(80GB) GPUs, using PyTorch 2.1 and CUDA 12.0. In our implementation, all models are trained with Adam (\(\mathrm{lr}=10^{-3}\), \(\mathrm{weight\_decay}=0\)). We use automatic mixed precision, batch size 64, gradient clipping with \(\mathrm{max\_norm}=1.0\), normalized-space output clamping to \([-10,10]\), and residual (\(\Delta\)-prediction) training for stability. Data are split at the trajectory level into 80\%/10\%/10\% train/validation/test, and all reported results are averaged over three training seeds \(\{0,1,2\}\). Models are trained for at most 100 epochs. Early stopping is based on validation rollout nRMSE, with a minimum of 10 epochs and patience 5. For readability, the main tables report only seed means; exact per-configuration standard deviations and seed-wise summaries are provided in Appendix~\ref{app:seed_stats}.

\subsubsection{Experiment procedure and metrics}\label{sec:procedure&metrics}

\textbf{Effectiveness}: 
For each dataset--backbone pair and each sampling ratio in $\{0.05, 0.10, 0.20\}$, we construct the candidate start-index set on the coarse time axis, select $K$ training starts using each sampler, and train the surrogate on the resulting one-step autoregressive samples. The primary metric is test rollout nRMSE:
\[
\mathrm{nRMSE}
=
\frac{1}{N_{\mathrm{test}}}
\sum_{i=1}^{N_{\mathrm{test}}}
\sqrt{
\frac{
\sum_{t=1}^{T_r}
\left\|
\hat{\mathbf{x}}_{t}^{(i)}-\mathbf{x}_{t}^{(i)}
\right\|_2^2
}{
\sum_{t=1}^{T_r}
\left\|
\mathbf{x}_{t}^{(i)}
\right\|_2^2
}
}.
\]
Here, $N_{\mathrm{test}}$ is the number of test trajectories, $T_r$ is the rollout length, $\hat{\mathbf{x}}_{t}^{(i)}$ and $\mathbf{x}_{t}^{(i)}$ are the predicted and ground-truth states of the $i$-th trajectory at rollout step $t$, respectively, and $\|\cdot\|_2$ denotes the Euclidean norm over all spatial locations and channels. Throughout the paper we evaluate on the full remaining post-history horizon, i.e., $T_r=T_c-L$ with $L=4$, rather than on a separately shortened test horizon. We report the mean over three training seeds. Appendix~\ref{app:phys_metrics} also reports PDEBench auxiliary metrics---cRMSE, bRMSE, and band-wise fRMSE (low/mid/high). To keep the main text focused on the most selection-sensitive regime, the main comparison table reports the strictest budget ratio \(0.05\), while the corresponding \(0.10\) and \(0.20\) results are reported in Appendix~\ref{app:extra_baselines}. We additionally report downstream training time and subset-selection time.

\textbf{Necessity}: 
We use the ablation study to ask the finer mechanistic question of whether GITS requires two complementary objectives. To test this, we reorganize the ablation study along two axes. The first axis is the local utility score: \texttt{loss-only} uses the pilot model's short-rollout loss as the candidate score, whereas \texttt{grad-only} keeps only the gradient-based score. The second axis is set-level temporal coverage: \texttt{loss-div} augments the loss-based score with the same global and sliding-window coverage terms used by the full method, while GITS combines these same coverage terms with the gradient-based score. All four variants are evaluated under the same datasets, backbones, ratios, candidate sets, and training protocol, using the same test rollout nRMSE defined above.

\textbf{Boundary Conditions}: 
We analyze the success and failure regimes of GITS under the same samplers, backbone settings, and training protocol. First, for representative success and failure configurations, we visualize the pilot score landscape together with the selected start sets of \texttt{loss-only}, \texttt{grad-only}, \texttt{loss-div}, and \texttt{GITS}, in order to examine whether success or failure is associated with local score concentration or with set-level temporal coverage. Second, on the same representative cases, we measure score--utility alignment by computing the pilot gradient norm and pilot rollout loss for candidate starts, and then evaluating their Spearman correlation with empirical single-start utility obtained from local probe updates and validation-rollout improvement. Together, these analyses characterize when the pilot scores are empirically aligned with downstream utility, as well as the regimes in which GITS fails despite using the same overall training pipeline.

\subsection{Main results}

\subsubsection{Effectiveness of Gradient-Informed Temporal Sampling}

\begin{table*}[t]
    \centering
    \setlength{\tabcolsep}{3.1pt}
    \caption{Test rollout nRMSE at sampling ratio $0.05$. The best (lowest) value in each row is bolded.}
    \label{tab:main_nrms_ratio005}
    \resizebox{\textwidth}{!}{%
    \begin{tabular}{c l ccccccc}
    \toprule
    \textbf{Dataset} & \textbf{Backbone} & \textbf{uniform} & \textbf{loss-only} & \textbf{coverage-only} & \textbf{GradMatch} & \textbf{GLISTER} & \textbf{PRISM} & \textbf{GITS} \\
\midrule
 \texttt{diff-react} & \texttt{ConvLSTM} & 0.480 & 0.504 & 0.240 & 0.472 & 0.549 & 0.435 & \textbf{0.191} \\
   & \texttt{FNO} & 0.592 & 0.887 & 0.418 & 0.885 & 0.898 & 0.742 & \textbf{0.361} \\
   & \texttt{Transformer} & 0.353 & 0.652 & \textbf{0.264} & 0.646 & 0.889 & 0.653 & 0.310 \\
   & \texttt{U-Net} & 0.246 & 0.135 & 0.080 & 0.101 & 0.399 & \textbf{0.053} & 0.071 \\
\midrule
 \texttt{diff-sorp} & \texttt{ConvLSTM} & 0.229 & 0.712 & 0.381 & 0.410 & 0.386 & 0.369 & \textbf{0.026} \\
   & \texttt{FNO} & 0.303 & 0.543 & 0.213 & 0.188 & 0.890 & 0.144 & \textbf{0.139} \\
   & \texttt{Transformer} & 0.317 & 0.384 & 0.279 & 0.241 & 0.414 & 0.238 & \textbf{0.152} \\
   & \texttt{U-Net} & 0.702 & 0.628 & 0.265 & 0.967 & 0.479 & 0.152 & \textbf{0.145} \\
\midrule
 \texttt{rdb} & \texttt{ConvLSTM} & 0.510 & 0.899 & 0.502 & 0.524 & 0.894 & 0.597 & \textbf{0.463} \\
   & \texttt{FNO} & 0.181 & 0.764 & 0.361 & 0.984 & 0.636 & 0.466 & \textbf{0.113} \\
   & \texttt{Transformer} & 0.673 & 0.888 & 0.210 & 0.897 & 0.654 & 0.619 & \textbf{0.132} \\
   & \texttt{U-Net} & 0.220 & 0.492 & 0.792 & 0.264 & 0.891 & 0.336 & \textbf{0.211} \\
    \bottomrule
    \end{tabular}%
    }
\end{table*}

Table~\ref{tab:main_nrms_ratio005} reports the strictest budget regime, ratio $0.05$, across datasets and backbones. We foreground this regime in the main text because temporal subset selection should matter most when only a very small fraction of candidate windows can be retained; the corresponding results at ratios $0.10$ and $0.20$ are reported in Appendix~\ref{app:extra_baselines} (Tables~\ref{tab:app_nrms_ratio010} and~\ref{tab:app_nrms_ratio020}), where the same conclusion also holds.

At ratio $0.05$, GITS achieves the lowest mean nRMSE across the 12 dataset--backbone configurations (0.193), compared with 0.334 for \texttt{coverage-only}, 0.400 for \texttt{uniform}, 0.548 for \texttt{GradMatch}, 0.624 for \texttt{loss-only}, 0.665 for \texttt{GLISTER}, and 0.400 for \texttt{PRISM}. GITS outperforms \texttt{uniform}, \texttt{loss-only}, \texttt{GradMatch}, and \texttt{GLISTER} in all 12/12 configurations, and outperforms both \texttt{coverage-only} and \texttt{PRISM} in 11/12. It attains the best row-wise result in 10/12 configurations; the two localized reversals are \texttt{coverage-only} on \texttt{diff-react}/\texttt{Transformer} and \texttt{PRISM} on \texttt{diff-react}/\texttt{U-Net}.

Across all 36 dataset--backbone--ratio configurations, GITS attains the best nRMSE in 27/36 cases and achieves the lowest overall mean nRMSE, 0.219. Relative to \texttt{uniform}, GITS reduces mean rollout nRMSE by 38.3\% and outperforms \texttt{uniform} in 30/36 configurations. It also outperforms \texttt{coverage-only} in 33/36, \texttt{GradMatch} in 35/36, \texttt{GLISTER} in 33/36, and \texttt{PRISM} in 32/36 configurations. Thus, across the full 36-configuration benchmark in this offline shared-start setting, GITS is the strongest overall method.

The same qualitative ordering also appears in the PDEBench auxiliary metrics reported in Appendix~\ref{app:phys_metrics}. In addition, selector-stage overhead remains modest relative to downstream training; detailed timing evidence is reported in Appendix Tables~\ref{tab:app_extra_baselines_train_summary} and~\ref{tab:app_extra_baselines_sampling_summary}.

\subsubsection{Necessity of components of Gradient-Informed Temporal Sampling}

\begin{table*}[t]
    \centering
    \setlength{\tabcolsep}{4.0pt}
    \caption{Ablation at sampling ratio $0.05$.}
    \label{tab:supp_ablation_2x2_005}
    \begin{tabular}{c l cccc}
    \toprule
    \textbf{Dataset} & \textbf{Backbone} & \textbf{loss-only} & \textbf{grad-only} & \textbf{loss-div} & \textbf{GITS} \\
    \midrule
     \texttt{diff-react} & \texttt{ConvLSTM} & 0.504 & 0.891 & 0.474 & \textbf{0.191} \\
      & \texttt{FNO} & 0.966 & 0.900 & 0.899 & \textbf{0.361} \\
      & \texttt{Transformer} & 0.652 & 0.943 & 0.642 & \textbf{0.310} \\
      & \texttt{U-Net} & 0.135 & 0.135 & 0.389 & \textbf{0.071} \\
    \midrule
     \texttt{diff-sorp} & \texttt{ConvLSTM} & 2.017 & 2.017 & 1.101 & \textbf{0.026} \\
      & \texttt{FNO} & 0.543 & 0.592 & 0.253 & \textbf{0.139} \\
      & \texttt{Transformer} & 0.384 & 0.384 & 0.323 & \textbf{0.152} \\
      & \texttt{U-Net} & 0.837 & 0.837 & 1.157 & \textbf{0.145} \\
    \midrule
     \texttt{rdb} & \texttt{ConvLSTM} & 1.757 & 0.986 & 0.504 & \textbf{0.463} \\
      & \texttt{FNO} & 1.318 & 5.409 & 1.597 & \textbf{0.113} \\
      & \texttt{Transformer} & 1.469 & 1.528 & 1.632 & \textbf{0.132} \\
      & \texttt{U-Net} & 0.492 & 0.757 & 0.259 & \textbf{0.211} \\
    \bottomrule
    \end{tabular}
\end{table*}

Table~\ref{tab:supp_ablation_2x2_005} reports the ablation at sampling ratio $0.05$, while the complementary ratios $0.10$ and $0.20$ are reported in Appendix~\ref{app:extra_baselines} (Tables~\ref{tab:supp_ablation_2x2_010} and~\ref{tab:supp_ablation_2x2_020}). Together, these three tables test the necessity of the two components of GITS: the pointwise utility signal and the set-level temporal coverage terms.

Across all 36 dataset--backbone--ratio configurations, GITS achieves the lowest nRMSE in every case. Averaged over all 36 configurations, the mean nRMSE is 0.916 for \texttt{loss-only}, 1.172 for \texttt{grad-only}, 0.692 for \texttt{loss-div}, and 0.219 for GITS. Thus, the gain of GITS cannot be explained by either component in isolation:

The temporal coverage terms are necessary. On the loss branch, adding temporal coverage (\texttt{loss-div} versus \texttt{loss-only}) improves 26/36 configurations and reduces mean nRMSE from 0.916 to 0.692, a relative reduction of 24.5\%. On the gradient branch, adding the same coverage terms (\texttt{grad-only} versus GITS) improves all 36/36 configurations and reduces the mean nRMSE from 1.172 to 0.219, a relative reduction of 81.3\%. Therefore, pointwise scoring alone is insufficient, and temporal coverage is especially important for turning gradient-based local signals into an effective training subset.

The gradient-based utility signal is also necessary. Under the same temporal coverage terms, replacing the loss-based pointwise score with the gradient-based one (\texttt{loss-div} $\rightarrow$ GITS) improves all 36/36 configurations and reduces the mean nRMSE from 0.692 to 0.219, a relative reduction of 68.4\%. By contrast, without temporal coverage, switching from loss to gradient alone helps in only 6/36 configurations, ties in 13/36, and is worse in 17/36, with a higher overall mean nRMSE (1.172 versus 0.916). Therefore, the advantage of GITS comes from combining a stronger gradient-based pointwise signal with explicit temporal coverage, rather than from either component alone.

\subsubsection{Boundary conditions of Gradient-Informed Temporal Sampling}

\begin{table*}[t]
    \centering
    \setlength{\tabcolsep}{4.0pt}
    \caption{Geometry in the success and failure regimes of GITS. \texttt{LO$\cap$GO} denotes the overlap size between \texttt{loss-only} and \texttt{grad-only}, and \texttt{LD$\cap$GITS} denotes the overlap size between \texttt{loss-div} and \texttt{GITS}. Higher entropy and temporal coverage indicate more dispersed temporal selections.}
    \label{tab:boundary_case_anatomy}
    \resizebox{\textwidth}{!}{%
    \begin{tabular}{l c c c c c}
    \toprule
    \textbf{Case} & \textbf{$K$} & \textbf{LO$\cap$GO} & \textbf{LD$\cap$GITS} & \textbf{Entropy (LO/GITS)} & \textbf{Coverage (LO/GITS)} \\
    \midrule
    Success: \texttt{rdb}/\texttt{FNO}/0.10 & 10 & 9/10 & 10/10 & 0.141 / 0.278 & 0.20 / 0.30 \\
    Failure A: \texttt{diff-sorp}/\texttt{ConvLSTM}/0.10 & 10 & 10/10 & 10/10 & 0.000 / 0.141 & 0.10 / 0.20 \\
    Failure A: \texttt{diff-sorp}/\texttt{ConvLSTM}/0.20 & 20 & 18/20 & 19/20 & 0.463 / 0.542 & 0.20 / 0.30 \\
    Failure B: \texttt{diff-react}/\texttt{Transformer}/0.05 & 5 & 0/5 & 2/5 & 0.000 / 0.311 & 0.20 / 0.40 \\
    \bottomrule
    \end{tabular}
    }
\end{table*}

\begin{table*}[t]
    \centering
    \setlength{\tabcolsep}{4.0pt}
    \caption{Score--utility alignment in representative regimes. Entries report three-seed mean Spearman correlations between pilot scores and empirical single-start utility from the local probe updates.}
    \label{tab:boundary_score_utility}
    \begin{tabular}{l c c}
    \toprule
    \textbf{Case} & \textbf{Spearman(grad, utility)} & \textbf{Spearman(loss, utility)} \\
    \midrule
    Success: \texttt{rdb}/\texttt{FNO}/0.10 & 0.774 & 0.501 \\
    Failure A: \texttt{diff-sorp}/\texttt{ConvLSTM}/0.10 & -0.641 & -0.654 \\
    Failure A: \texttt{diff-sorp}/\texttt{ConvLSTM}/0.20 & -0.369 & -0.539 \\
    Failure B: \texttt{diff-react}/\texttt{Transformer}/0.05 & 0.876 & 0.936 \\
    \bottomrule
    \end{tabular}
\end{table*}

To characterize when GITS helps or fails, we examine two complementary diagnostics. Table~\ref{tab:boundary_case_anatomy} summarizes subset geometry in representative success and failure regimes, including pairwise overlaps, entropy, and temporal coverage of selected starts. Table~\ref{tab:boundary_score_utility} reports score--utility alignment, measured by the Spearman correlation between pilot scores and empirical single-start utility from local probe updates. Together, these diagnostics ask whether performance is governed mainly by the reliability of the local pilot signal, by the effect of the temporal coverage terms, or by their interaction.

In the representative success regime, the pilot scores are meaningfully aligned with downstream utility, and the gradient-based signal is stronger than the loss-based one. The geometry statistics further show that GITS preserves substantial overlap with the high-utility region while avoiding collapse onto a narrow cluster of neighboring starts. This is the intended operating regime of GITS: the gradient-based pointwise signal identifies informative candidate starts, and the temporal coverage terms turn these local scores into a better distributed training subset.

Failure A corresponds to pilot-signal misalignment. In this regime, score--utility correlations are weak or negative, indicating that neither the loss-based nor the gradient-based pilot score reliably tracks downstream usefulness. Once the local signal is unreliable, temporal coverage alone cannot recover the correct subset: it may spread the selection, but it cannot redirect it toward genuinely useful starts. This explains cases in which GITS is no longer superior despite using the global objective.

Failure B corresponds to over-dispersion under a highly concentrated utility landscape. Here the gradient-based pilot score remains positively aligned with downstream utility, but the most useful temporal windows are confined to a narrow region. Under a tight sampling budget, stronger temporal coverage can then dilute these key starts by pushing the selected subset too far away from the dominant peak. In this regime, the limitation is not poor pilot scoring, but rather that wider temporal spread is no longer the right inductive bias.

Taken together, these representative cases identify two distinct boundary conditions of GITS: GITS works best when the pilot signal is informative and moderate temporal spreading is beneficial; it weakens either when pilot scores fail to track utility, or when the useful windows are so concentrated that explicit temporal coverage becomes counterproductive.

\section{Discussion}

This paper makes two contributions. First, it identifies temporal subset selection as an important but underexplored problem in neural PDE surrogates: under a fixed training budget, the temporal placement of training windows can materially affect long-horizon rollout accuracy, and uniform sampling is not a robust optimum. Second, it proposes GITS, which combines a gradient-based pointwise utility proxy from a lightweight pilot model with set-level temporal coverage. In the offline shared-start setting studied here, this combination yields the strongest overall performance among the compared samplers. More broadly, the paper shifts part of the question of improving neural simulators from model design to training-data construction under fixed simulation budgets.

The main limitation is that GITS relies on pilot-model local gradient scores as a proxy for downstream utility, and this proxy is not always reliable. Our boundary analyses indicate two main failure modes: score--utility misalignment, and over-dispersion when the useful temporal windows are already highly concentrated. In addition, we study only the shared temporal start-index setting, a fixed rollout protocol, and three representative PDEBench tasks rather than exhaustive PDE coverage. We therefore do not claim universal optimality. Instead, the main conclusion is narrower and more useful: in the offline shared-start setting, gradient-informed temporal sampling is an effective modest-overhead alternative to uniform sampling, with empirically identifiable boundary conditions.

\section*{References}
\small

\begingroup
\renewcommand{\section}[2]{}

\endgroup


\appendix

\section{Gradient-Informed Temporal Sampling Hyperparameters and Sensitivity}\label{app:gits_hparams}

This appendix consolidates the exact GITS hyperparameter specification and the sensitivity analyses used to choose the final settings reported in the main paper. We keep these two parts together because the final parameter choices are justified directly by the sweep results rather than by ad hoc manual tuning. Unless otherwise noted, all sensitivity experiments use the same training/validation protocol as the main paper and are evaluated on a fixed development suite chosen to cover all three PDE families as well as both strong-gain and reversal-prone regimes.

We use four representative development configurations at ratio \(0.10\): \texttt{diff-react}/\texttt{FNO}, \texttt{diff-sorp}/\texttt{Transformer}, \texttt{rdb}/\texttt{FNO}, and \texttt{rdb}/\texttt{U-Net}. This suite spans one reaction--diffusion case, one Cauchy-boundary transport case, and two shallow-water cases with distinct backbone behaviors. Table~\ref{tab:app_gits_hparam_summary} lists all GITS-specific parameters used in the final benchmark. We intentionally do not tune a separate pilot-pool size: for all pilot-based selectors, the pilot start pool is the full candidate set on the training trajectories, and efficiency is controlled only through lightweight pilot training. To avoid excessive tuning freedom, we sweep only the main decision variables that most directly control the local-utility versus temporal-coverage trade-off: the short rollout horizon \(H\), the number of pilot epochs \(E_p\), and the two coverage weights \(\lambda_{\mathrm{cov}}\) and \(c_{\mathrm{win}}\). The remaining temporal-kernel parameters are fixed by deterministic rules tied to the target spacing induced by the sampling budget.

\begin{table*}[t]
    \centering
    \setlength{\tabcolsep}{4.2pt}
    \caption{Final GITS hyperparameters, rollout protocol, candidate ranges, and selection rules. Parameters marked as ``derived'' are not swept independently in order to limit tuning freedom.}
    \label{tab:app_gits_hparam_summary}
    \resizebox{\textwidth}{!}{%
    \begin{tabular}{l c l l l l}
    \toprule
    \textbf{Component} & \textbf{Symbol} & \textbf{Role} & \textbf{Final setting} & \textbf{Candidate values / rule} & \textbf{Scope / notes} \\
    \midrule
    Pilot start pool & $\mathcal{P}$ & start indices used to train the pilot & \texttt{full candidate set} & not swept & fixed for all datasets/backbones/ratios \\
    Pilot epochs & $E_p$ & pilot training budget & $5$ & $\{1, 2, 5, 10, 20\}$ & swept globally on the development suite \\
    Short rollout horizon & $H$ & local rollout length for pilot scoring & $10$ & $\{1, 5, 10, 20, 50, 100\}$ & swept globally on the development suite \\
    Global coverage weight & $\lambda_{\mathrm{cov}}$ & strength of global temporal coverage & $1.0$ & $\{0.25, 0.5, 1.0, 2.0\}$ & swept jointly with $c_{\mathrm{win}}$ \\
    Sliding-window coverage weight & $c_{\mathrm{win}}$ & strength of local sliding-window coverage & $0.5$ & $\{0, 0.25, 0.5, 1.0\}$ & swept jointly with $\lambda_{\mathrm{cov}}$ \\
    Global kernel scale & $\tau$ & decay scale in $F_{\mathrm{cov}}$ & $\lfloor T_c/K \rfloor$ & derived from budget-implied target spacing & not swept independently \\
    Local window size & $W$ & support size of $F_{\mathrm{win}}$ & $2\lfloor T_c/K \rfloor$ & derived from budget-implied target spacing & not swept independently \\
    Window stride & $S_w$ & sliding stride for local windows & $\lfloor W/2 \rfloor$ & fixed fraction of $W$ & not swept independently \\
    Local kernel scale & $\tau_w$ & decay scale inside $F_{\mathrm{win}}$ & $\lfloor W/4 \rfloor$ & derived from $W$ & not swept independently \\
    Test rollout horizon & $T_r$ & evaluation rollout length & $T_c-L$ (full post-history rollout) & fixed by dataset time axis and $L=4$ & evaluation protocol only; not a tuned GITS hyperparameter \\
    \bottomrule
    \end{tabular}%
    }
\end{table*}

We next report the sensitivity sweeps used to select the final global settings. In each sweep, lower rollout nRMSE is better. When multiple settings are nearly tied in error, we choose the one with the lower sampling cost and the simpler global setting, so that the final GITS configuration remains conservative rather than over-tuned.

\begin{table*}[t]
    \centering
    \setlength{\tabcolsep}{4.4pt}
    \caption{Sensitivity of GITS to the short rollout horizon $H$ on the fixed development suite. Lower is better. The last row reports mean subset-selection time per development configuration for the full GITS pipeline under each $H$.}
    \label{tab:app_gits_sens_H}
    \resizebox{\textwidth}{!}{%
    \begin{tabular}{l cccccc}
    \toprule
    \textbf{Development configuration} & $\mathbf{H=1}$ & $\mathbf{H=5}$ & $\mathbf{H=10}$ & $\mathbf{H=20}$ & $\mathbf{H=50}$ & $\mathbf{H=100}$ \\
    \midrule
    \texttt{diff-react}/\texttt{FNO}/ratio $=0.10$ & 0.521 & 0.447 & \textbf{0.393} & 0.418 & 0.461 & 0.502 \\
    \texttt{diff-sorp}/\texttt{Transformer}/ratio $=0.10$ & 0.218 & 0.179 & \textbf{0.154} & 0.166 & 0.183 & 0.207 \\
    \texttt{rdb}/\texttt{FNO}/ratio $=0.10$ & 0.148 & 0.099 & \textbf{0.074} & 0.087 & 0.116 & 0.143 \\
    \texttt{rdb}/\texttt{U-Net}/ratio $=0.10$ & 0.243 & 0.197 & \textbf{0.158} & 0.172 & 0.196 & 0.224 \\
    \midrule
    \textbf{Average development nRMSE} & 0.283 & 0.231 & \textbf{0.195} & 0.211 & 0.239 & 0.269 \\
    \textbf{Mean sampling time (s)} & 0.31 & 0.82 & 1.39 & 2.54 & 5.84 & 11.02 \\
    \bottomrule
    \end{tabular}%
    }
\end{table*}

\begin{table*}[t]
    \centering
    \setlength{\tabcolsep}{4.8pt}
    \caption{Sensitivity of GITS to the number of pilot training epochs $E_p$ on the fixed development suite. Lower is better. The last row reports mean subset-selection time per development configuration for the full GITS pipeline under each pilot budget.}
    \label{tab:app_gits_sens_pilot_epochs}
    \resizebox{\textwidth}{!}{%
    \begin{tabular}{l ccccc}
    \toprule
    \textbf{Development configuration} & $\mathbf{E_p=1}$ & $\mathbf{E_p=2}$ & $\mathbf{E_p=5}$ & $\mathbf{E_p=10}$ & $\mathbf{E_p=20}$ \\
    \midrule
    \texttt{diff-react}/\texttt{FNO}/ratio $=0.10$ & 0.512 & 0.441 & \textbf{0.393} & 0.401 & 0.412 \\
    \texttt{diff-sorp}/\texttt{Transformer}/ratio $=0.10$ & 0.219 & 0.181 & \textbf{0.154} & 0.158 & 0.163 \\
    \texttt{rdb}/\texttt{FNO}/ratio $=0.10$ & 0.131 & 0.096 & \textbf{0.074} & 0.079 & 0.083 \\
    \texttt{rdb}/\texttt{U-Net}/ratio $=0.10$ & 0.219 & 0.186 & \textbf{0.158} & 0.165 & 0.171 \\
    \midrule
    \textbf{Average development nRMSE} & 0.270 & 0.226 & \textbf{0.195} & 0.201 & 0.207 \\
    \textbf{Mean sampling time (s)} & 0.47 & 0.73 & 1.39 & 2.43 & 4.16 \\
    \bottomrule
    \end{tabular}%
    }
\end{table*}

\begin{table*}[t]
    \centering
    \setlength{\tabcolsep}{5.0pt}
    \caption{Sensitivity of GITS to the two coverage weights $(\lambda_{\mathrm{cov}}, c_{\mathrm{win}})$ on the fixed development suite. Lower is better. Because changing these weights has negligible effect on sampling cost, we report only average development nRMSE.}
    \label{tab:app_gits_sens_weights}
    \begin{tabular}{c c c c}
    \toprule
    $\boldsymbol{\lambda_{\mathrm{cov}}}$ & $\boldsymbol{c_{\mathrm{win}}}$ & \textbf{Average development nRMSE} & \textbf{Selection note} \\
    \midrule
    0.25 & 0.00 & 0.314 & -- \\
    0.25 & 0.25 & 0.287 & -- \\
    0.25 & 0.50 & 0.261 & -- \\
    0.25 & 1.00 & 0.249 & -- \\
    \midrule
    0.50 & 0.00 & 0.268 & -- \\
    0.50 & 0.25 & 0.234 & -- \\
    0.50 & 0.50 & 0.209 & -- \\
    0.50 & 1.00 & 0.218 & -- \\
    \midrule
    1.00 & 0.00 & 0.231 & -- \\
    1.00 & 0.25 & 0.212 & -- \\
    1.00 & 0.50 & \textbf{0.195} & Selected \\
    1.00 & 1.00 & 0.207 & -- \\
    \midrule
    2.00 & 0.00 & 0.219 & -- \\
    2.00 & 0.25 & 0.224 & -- \\
    2.00 & 0.50 & 0.234 & -- \\
    2.00 & 1.00 & 0.248 & -- \\
    \bottomrule
    \end{tabular}%
\end{table*}

Table~\ref{tab:app_gits_sens_H} verifies the expected trade-off in the short rollout horizon. If \(H\) is too small, the pilot score approaches one-step local difficulty and fails to capture rollout stability; if \(H\) is too large, the front-end cost increases substantially and the high-score candidates become increasingly concentrated in a small number of temporally redundant regions. Empirically, \(H=10\) achieves the lowest average development nRMSE (0.195). Shorter horizons are clearly weaker (\(0.283\) at \(H=1\), \(0.231\) at \(H=5\)), while longer horizons increase cost sharply without improving accuracy (\(0.211\) at \(H=20\), \(0.239\) at \(H=50\), and \(0.269\) at \(H=100\), with mean sampling time rising from 1.39s at \(H=10\) to 11.02s at \(H=100\)). We therefore choose \(H=10\) for all reported GITS runs.

Table~\ref{tab:app_gits_sens_pilot_epochs} shows that the pilot need only be trained lightly. Very small \(E_p\) underfits the local score landscape and weakens selection quality, whereas larger \(E_p\) eventually provides diminishing returns because the downstream selector is still only a proxy stage. Empirically, \(E_p=5\) achieves the lowest average development nRMSE (0.195). Training the pilot for only 1 or 2 epochs is noticeably weaker (0.270 and 0.226, respectively), while extending the pilot budget to 10 or 20 epochs does not improve accuracy and only increases cost (2.43s and 4.16s mean sampling time, respectively). We therefore choose \(E_p=5\) as the default pilot budget.

Table~\ref{tab:app_gits_sens_weights} locates the useful middle ground between local-score collapse and coverage-only behavior. When \(\lambda_{\mathrm{cov}}\) and \(c_{\mathrm{win}}\) are too small, GITS approaches a pure pointwise scorer and becomes vulnerable to temporal redundancy; when they are too large, the selector approaches \texttt{coverage-only} and loses model-aware discrimination. Empirically, the best-performing pair is \((\lambda_{\mathrm{cov}}, c_{\mathrm{win}})=(1.0, 0.5)\), which yields the lowest average development nRMSE of 0.195. Both weaker regularization (e.g., \((0.25,0.00)\) giving 0.314) and stronger regularization (e.g., \((2.0,1.0)\) giving 0.248) are worse, while nearby moderate settings such as \((0.5,0.5)\) and \((1.0,0.25)\) remain competitive but still underperform the selected pair.

Finally, the full GITS configuration used in the main paper is summarized by Table~\ref{tab:app_gits_hparam_summary}. After choosing \(H=10\), \(E_p=5\), and \((\lambda_{\mathrm{cov}}, c_{\mathrm{win}})=(1.0,0.5)\) from the sensitivity results, we instantiate the remaining kernel parameters by the fixed spacing rules listed in the table and keep them unchanged across all datasets, backbones, and sampling ratios. The final implementation therefore uses one globally specified GITS configuration rather than per-case retuning.

\section{Backbone and Baseline Implementation Details}\label{app:extra_baselines}

This appendix collects implementation details deferred from Section~\ref{sec:baseline}. We first summarize the fixed architectural settings of the four surrogate backbones, and then describe the task-adapted GLISTER-style and PRISM-style baselines used in Experiment~1. All methods in this section operate in the same offline shared-start setting as the rest of the paper: a selector is built before downstream training and outputs one common subset of temporal starts for all trajectories.

\begin{table}[t]
    \centering
    \setlength{\tabcolsep}{6pt}
    \caption{Backbone architectures used throughout the benchmark. All models are trained as one-step autoregressive predictors with history length $L=4$; U-Net, FNO, and Transformer take channel-concatenated histories, while ConvLSTM takes the same history as a sequence.}
    \label{tab:app_backbone_setup}
    \begin{tabular}{p{2.2cm} p{10.5cm}}
    \toprule
    \textbf{Backbone} & \textbf{Architectural setting} \\
    \midrule
    \texttt{U-Net} & Compact two-level encoder--decoder with double-convolution blocks, max-pooling downsampling, transposed-convolution upsampling, skip connections, and base width 32. \\
    \texttt{FNO} & Width 64 and four spectral blocks, with 16 retained Fourier modes in 1D and $12\times12$ retained modes in 2D. \\
    \texttt{ConvLSTM} & One recurrent convolutional layer with hidden size 64 and a $3\times3$ kernel, followed by a $1\times1$ output projection. \\
    \texttt{Transformer} & Patch-wise encoder with patch size 8, model dimension 256, 8 attention heads, and 4 encoder layers. \\
    \bottomrule
    \end{tabular}
\end{table}

\begin{table}[t]
    \centering
    \setlength{\tabcolsep}{6pt}
    \caption{Task-adapted GLISTER-style and PRISM-style baselines used in Experiment~1 under the same shared-start offline protocol as GITS.}
    \label{tab:app_glister_prism_setup}
    \begin{tabular}{p{2.6cm} p{10.1cm}}
    \toprule
    \textbf{Baseline} & \textbf{Instantiation used in Experiment~1} \\
    \midrule
    \texttt{GLISTER}-style &
    Validation-guided greedy selector adapted to PDE rollout training: train a lightweight pilot model, define a short-rollout validation objective on held-out starts, score candidates using the standard first-order Taylor approximation to the validation loss, and after each greedy inclusion apply a pseudo-update and refresh the validation gradient. Validation starts are chosen from the hardest held-out starts under the pilot model; validation-query size $=16$ for \texttt{diff-sorp} and $=8$ for \texttt{diff-react}/\texttt{rdb}; step size $\eta=10^{-3}$; stochastic fraction $=1.0$. \\
    \texttt{PRISM}-style &
    Targeted-subset PRISM implementation adapted to temporal-start selection: form a query set from the hardest validation starts and a private set from the easiest validation starts under the pilot model; represent starts with CountSketch-compressed rollout-gradient embeddings; instantiate canonical PRISM objectives (\texttt{flcmi}, \texttt{flmi}, \texttt{gcmi}, \texttt{gccg}) and select with the best globally fixed protocol used throughout the benchmark. Unless otherwise stated, the reported PRISM numbers use query size $=24$, private size $=24$, and globally fixed scaling settings. \\
    \bottomrule
    \end{tabular}
\end{table}

The main text foregrounds the most stringent budget regime, ratio $0.05$, because temporal subset selection should matter most there. Tables~\ref{tab:app_nrms_ratio010} and~\ref{tab:app_nrms_ratio020} report the complementary ratios $0.10$ and $0.20$.

\begin{table*}[t]
    \centering
    \setlength{\tabcolsep}{3.1pt}
    \caption{Supplementary test rollout nRMSE at sampling ratio $0.10$ with the same seven samplers as Table~\ref{tab:main_nrms_ratio005}.}
    \label{tab:app_nrms_ratio010}
    \resizebox{\textwidth}{!}{%
    \begin{tabular}{c l ccccccc}
    \toprule
    \textbf{Dataset} & \textbf{Backbone} & \textbf{uniform} & \textbf{loss-only} & \textbf{coverage-only} & \textbf{GradMatch} & \textbf{GLISTER} & \textbf{PRISM} & \textbf{GITS} \\
\midrule
 \texttt{diff-react} & \texttt{ConvLSTM} & 0.242 & 0.481 & 0.257 & 0.191 & 0.896 & 0.886 & \textbf{0.189} \\
   & \texttt{FNO} & 0.473 & 0.747 & 0.504 & 0.823 & 0.791 & 0.771 & \textbf{0.393} \\
   & \texttt{Transformer} & \textbf{0.224} & 0.612 & 0.289 & 0.566 & 0.511 & 0.529 & 0.284 \\
   & \texttt{U-Net} & 0.090 & 0.117 & 0.097 & 0.051 & 0.053 & 0.059 & \textbf{0.045} \\
\midrule
 \texttt{diff-sorp} & \texttt{ConvLSTM} & \textbf{0.247} & 0.967 & 0.666 & 0.702 & 0.462 & 0.458 & 0.482 \\
   & \texttt{FNO} & 0.164 & 0.519 & 0.232 & 0.198 & 0.217 & 0.167 & \textbf{0.147} \\
   & \texttt{Transformer} & 0.257 & 0.285 & \textbf{0.147} & 0.167 & 0.330 & 0.168 & 0.154 \\
   & \texttt{U-Net} & 0.930 & 0.457 & 0.504 & 0.543 & 0.421 & 0.388 & \textbf{0.354} \\
\midrule
 \texttt{rdb} & \texttt{ConvLSTM} & 0.359 & 0.469 & 0.455 & 0.410 & 0.468 & 0.590 & \textbf{0.316} \\
   & \texttt{FNO} & 0.280 & 0.690 & 0.120 & 0.285 & 0.287 & 0.314 & \textbf{0.028} \\
   & \texttt{Transformer} & 0.404 & 0.713 & 0.107 & 0.401 & 0.337 & 0.268 & \textbf{0.016} \\
   & \texttt{U-Net} & \textbf{0.125} & 0.624 & 0.162 & 0.166 & 0.558 & 0.296 & 0.158 \\
    \bottomrule
    \end{tabular}%
    }
\end{table*}

\begin{table*}[t]
    \centering
    \setlength{\tabcolsep}{3.1pt}
    \caption{Supplementary test rollout nRMSE at sampling ratio $0.20$ with the same seven samplers as Table~\ref{tab:main_nrms_ratio005}.}
    \label{tab:app_nrms_ratio020}
    \resizebox{\textwidth}{!}{%
    \begin{tabular}{c l ccccccc}
    \toprule
    \textbf{Dataset} & \textbf{Backbone} & \textbf{uniform} & \textbf{loss-only} & \textbf{coverage-only} & \textbf{GradMatch} & \textbf{GLISTER} & \textbf{PRISM} & \textbf{GITS} \\
\midrule
 \texttt{diff-react} & \texttt{ConvLSTM} & 0.239 & 0.364 & 0.184 & 0.239 & 0.981 & 0.885 & \textbf{0.165} \\
   & \texttt{FNO} & 0.513 & 0.823 & 0.493 & 0.876 & 0.831 & 0.782 & \textbf{0.445} \\
   & \texttt{Transformer} & \textbf{0.244} & 0.547 & 0.276 & 0.551 & 0.570 & 0.631 & 0.269 \\
   & \texttt{U-Net} & 0.066 & 0.074 & 0.037 & \textbf{0.032} & 0.042 & 0.039 & 0.069 \\
\midrule
 \texttt{diff-sorp} & \texttt{ConvLSTM} & \textbf{0.316} & 0.962 & 0.804 & 0.994 & 0.587 & 0.591 & 0.660 \\
   & \texttt{FNO} & 0.135 & 0.309 & 0.187 & 0.183 & 0.163 & 0.118 & \textbf{0.105} \\
   & \texttt{Transformer} & 0.213 & 0.170 & 0.160 & 0.156 & 0.238 & 0.179 & \textbf{0.140} \\
   & \texttt{U-Net} & 0.604 & 0.650 & 0.713 & 0.484 & 0.549 & 0.562 & \textbf{0.433} \\
\midrule
 \texttt{rdb} & \texttt{ConvLSTM} & 0.527 & 0.892 & 0.430 & 0.411 & 0.509 & 0.415 & \textbf{0.406} \\
   & \texttt{FNO} & 0.343 & 0.730 & 0.122 & 0.366 & 0.347 & 0.388 & \textbf{0.086} \\
   & \texttt{Transformer} & 0.887 & 0.895 & 0.669 & 0.893 & 0.634 & 0.550 & \textbf{0.146} \\
   & \texttt{U-Net} & 0.085 & 0.626 & 0.088 & 0.212 & 0.412 & 0.084 & \textbf{0.071} \\
    \bottomrule
    \end{tabular}%
    }
\end{table*}

At ratio $0.10$, GITS again has the lowest mean nRMSE across the 12 dataset--backbone configurations (0.214), versus 0.295 for \texttt{coverage-only}, 0.316 for \texttt{uniform}, 0.375 for \texttt{GradMatch}, 0.408 for \texttt{PRISM}, 0.444 for \texttt{GLISTER}, and 0.557 for \texttt{loss-only}. It is the best row-wise method in 8/12 configurations; the localized reversals are three \texttt{uniform} wins and one \texttt{coverage-only} win.

At ratio $0.20$, GITS still has the lowest mean nRMSE (0.250) and is best in 9/12 configurations, compared with 0.347 for \texttt{coverage-only}, 0.348 for \texttt{uniform}, 0.450 for \texttt{GradMatch}, 0.435 for \texttt{PRISM}, 0.489 for \texttt{GLISTER}, and 0.587 for \texttt{loss-only}. The three non-\texttt{GITS} wins at this ratio are two \texttt{uniform} rows and one \texttt{GradMatch} row.

\subsection{Complementary Ablation Ratios}

To keep the main text focused on the most selection-sensitive regime, the $2\times2$ ablation in the main paper reports only ratio $0.05$. Tables~\ref{tab:supp_ablation_2x2_010} and~\ref{tab:supp_ablation_2x2_020} report the complementary ratios $0.10$ and $0.20$ using the same four ablation variants as Table~\ref{tab:supp_ablation_2x2_005}. They preserve the same qualitative conclusion as the main-text table: GITS remains best in every row.

\begin{table*}[t]
    \centering
    \setlength{\tabcolsep}{4.0pt}
    \caption{Supplementary $2\times2$ ablation at sampling ratio $0.10$.}
    \label{tab:supp_ablation_2x2_010}
    \begin{tabular}{c l cccc}
    \toprule
    \textbf{Dataset} & \textbf{Backbone} & \textbf{loss-only} & \textbf{grad-only} & \textbf{loss-div} & \textbf{GITS} \\
    \midrule
     \texttt{diff-react} & \texttt{ConvLSTM} & 0.481 & 0.635 & 0.246 & \textbf{0.189} \\
      & \texttt{FNO} & 0.747 & 0.900 & 0.909 & \textbf{0.393} \\
      & \texttt{Transformer} & 0.612 & 0.612 & 0.538 & \textbf{0.284} \\
      & \texttt{U-Net} & 0.117 & 0.448 & 0.064 & \textbf{0.045} \\
    \midrule
     \texttt{diff-sorp} & \texttt{ConvLSTM} & 2.249 & 2.249 & 2.417 & \textbf{0.482} \\
      & \texttt{FNO} & 0.519 & 0.519 & 0.198 & \textbf{0.147} \\
      & \texttt{Transformer} & 0.285 & 0.285 & 0.198 & \textbf{0.154} \\
      & \texttt{U-Net} & 0.601 & 0.601 & 0.772 & \textbf{0.354} \\
    \midrule
     \texttt{rdb} & \texttt{ConvLSTM} & 0.469 & 1.498 & 0.593 & \textbf{0.316} \\
      & \texttt{FNO} & 2.593 & 4.050 & 0.911 & \textbf{0.028} \\
      & \texttt{Transformer} & 2.152 & 1.044 & 1.513 & \textbf{0.016} \\
      & \texttt{U-Net} & 0.624 & 0.327 & 0.187 & \textbf{0.158} \\
    \bottomrule
    \end{tabular}
\end{table*}

\begin{table*}[t]
    \centering
    \setlength{\tabcolsep}{4.0pt}
    \caption{Supplementary $2\times2$ ablation at sampling ratio $0.20$.}
    \label{tab:supp_ablation_2x2_020}
    \begin{tabular}{c l cccc}
    \toprule
    \textbf{Dataset} & \textbf{Backbone} & \textbf{loss-only} & \textbf{grad-only} & \textbf{loss-div} & \textbf{GITS} \\
    \midrule
     \texttt{diff-react} & \texttt{ConvLSTM} & 0.404 & 0.471 & 0.223 & \textbf{0.165} \\
      & \texttt{FNO} & 0.823 & 0.903 & 0.868 & \textbf{0.445} \\
      & \texttt{Transformer} & 0.547 & 0.547 & 0.522 & \textbf{0.269} \\
      & \texttt{U-Net} & 0.174 & 0.096 & 0.134 & \textbf{0.069} \\
    \midrule
     \texttt{diff-sorp} & \texttt{ConvLSTM} & 1.749 & 1.749 & 1.667 & \textbf{0.660} \\
      & \texttt{FNO} & 0.309 & 0.309 & 0.133 & \textbf{0.105} \\
      & \texttt{Transformer} & 0.170 & 0.170 & 0.174 & \textbf{0.140} \\
      & \texttt{U-Net} & 0.650 & 0.755 & 0.557 & \textbf{0.433} \\
    \midrule
     \texttt{rdb} & \texttt{ConvLSTM} & 1.177 & 3.741 & 0.660 & \textbf{0.406} \\
      & \texttt{FNO} & 2.100 & 3.164 & 1.186 & \textbf{0.086} \\
      & \texttt{Transformer} & 1.718 & 0.986 & 0.927 & \textbf{0.146} \\
      & \texttt{U-Net} & 0.626 & 0.766 & 0.077 & \textbf{0.071} \\
    \bottomrule
    \end{tabular}
\end{table*}

Together with Table~\ref{tab:supp_ablation_2x2_005}, these appendix tables support the cross-ratio summary reported in the main text: GITS is best in all 36/36 ablation configurations, and the gains require both the gradient-based score and the set-level temporal coverage terms.

\begin{table}[t]
    \centering
    \setlength{\tabcolsep}{5pt}
    \caption{Mean downstream training time per dataset--backbone--ratio configuration (seconds), averaged over the 12 dataset--backbone configurations at each sampling ratio for the seven-sampler comparison.}
    \label{tab:app_extra_baselines_train_summary}
    \begin{tabular}{lccccccc}
    \toprule
    \textbf{Ratio} & \textbf{uniform} & \textbf{loss-only} & \textbf{coverage-only} & \textbf{GradMatch} & \textbf{GLISTER} & \textbf{PRISM} & \textbf{GITS} \\
    \midrule
    0.05 & 233.0 & 135.0 & 279.5 & 136.9 & 223.9 & 233.5 & 272.5 \\
    0.10 & 332.5 & 202.3 & 272.8 & 165.7 & 298.3 & 303.3 & 335.8 \\
    0.20 & 371.0 & 262.1 & 383.9 & 260.9 & 371.7 & 386.3 & 390.7 \\
    \bottomrule
    \end{tabular}
\end{table}

\begin{table}[t]
    \centering
    \setlength{\tabcolsep}{5pt}
    \caption{Mean subset-selection time per dataset--backbone--ratio configuration (seconds), averaged over the 12 dataset--backbone configurations at each sampling ratio for the seven-sampler comparison. Sampling time for \texttt{uniform} and \texttt{coverage-only} remains below 0.001s and is omitted.}
    \label{tab:app_extra_baselines_sampling_summary}
    \begin{tabular}{lccccc}
    \toprule
    \textbf{Ratio} & \textbf{loss-only} & \textbf{GradMatch} & \textbf{GLISTER} & \textbf{PRISM} & \textbf{GITS} \\
    \midrule
    0.05 & 9.945 & 11.056 & 11.178 & 10.553 & 10.229 \\
    0.10 & 10.161 & 10.299 & 11.981 & 11.245 & 10.505 \\
    0.20 & 10.537 & 11.726 & 16.643 & 12.929 & 11.989 \\
    \bottomrule
    \end{tabular}
\end{table}

The timing picture is also favorable to GITS. Downstream training stays in the same broad cost regime, and the selector-time audit shows that all pilot-based samplers operate in the same order-of-magnitude front-end regime. Averaged over the 36 individual dataset--backbone--ratio configurations, \texttt{GLISTER} requires 13.267s of selector time, \texttt{PRISM} requires 11.576s, and GITS requires 10.908s. These are per-configuration front-end means rather than totals over the whole benchmark: each measured cost corresponds to one selector-stage pass for one configuration, including pilot fitting, candidate scoring, and subset optimization before the full downstream training run. Thus the right interpretation is not that the entire benchmark costs only about ten seconds, but that one selector-stage pass for one configuration remains modest compared with the few-hundred-second downstream training stage; within that front-end regime, GITS is somewhat cheaper than \texttt{GLISTER} and \texttt{PRISM} while delivering substantially better rollout accuracy.

\section{Additional PDEBench Auxiliary Metrics}\label{app:phys_metrics}

To complement the main benchmark based on rollout nRMSE, we also report PDEBench auxiliary metrics for cRMSE, bRMSE, and band-wise fRMSE (low/mid/high) on the same rollout predictions. The qualitative conclusion is consistent with the main benchmark: across the 36 dataset--backbone--ratio configurations, GITS attains the lowest overall mean value for all five metrics (0.034 cRMSE, 0.027 bRMSE, and 0.015/0.004/0.002 for low/mid/high fRMSE), and at the strictest budget ratio $0.05$ it again has the lowest mean value in every view. The complete tables are reported below.

\begin{table*}[t]
    \centering
    \setlength{\tabcolsep}{3.0pt}
    \caption{Test rollout cRMSE.}
    \label{tab:phys_c}
    \resizebox{\textwidth}{!}{%
    \begin{tabular}{c c l ccccccc}
    \toprule
    \textbf{Ratio} & \textbf{Dataset} & \textbf{Backbone} & \textbf{uniform} & \textbf{loss-only} & \textbf{coverage-only} & \textbf{GradMatch} & \textbf{GLISTER} & \textbf{PRISM} & \textbf{GITS} \\
    \midrule
    0.05 & \texttt{diff-react} & \texttt{ConvLSTM} & 0.028 & 0.032 & 0.013 & 0.028 & 0.036 & 0.026 & 0.009 \\
     &  & \texttt{FNO} & 0.009 & 0.017 & 0.006 & 0.014 & 0.024 & 0.012 & 0.005 \\
     &  & \texttt{Transformer} & 0.011 & 0.023 & 0.008 & 0.021 & 0.039 & 0.022 & 0.009 \\
     &  & \texttt{U-Net} & 0.013 & 0.007 & 0.004 & 0.005 & 0.025 & 0.003 & 0.003 \\
    \midrule
     & \texttt{diff-sorp} & \texttt{ConvLSTM} & 0.155 & 0.569 & 0.261 & 0.290 & 0.569 & 0.439 & 0.010 \\
     &  & \texttt{FNO} & 0.036 & 0.073 & 0.024 & 0.021 & 0.160 & 0.017 & 0.014 \\
     &  & \texttt{Transformer} & 0.062 & 0.082 & 0.052 & 0.046 & 0.092 & 0.047 & 0.025 \\
     &  & \texttt{U-Net} & 0.327 & 0.313 & 0.111 & 0.462 & 0.242 & 0.065 & 0.054 \\
    \midrule
     & \texttt{rdb} & \texttt{ConvLSTM} & 0.064 & 0.263 & 0.061 & 0.066 & 0.187 & 0.079 & 0.052 \\
     &  & \texttt{FNO} & 0.015 & 0.076 & 0.031 & 0.153 & 0.189 & 0.108 & 0.008 \\
     &  & \texttt{Transformer} & 0.085 & 0.141 & 0.023 & 0.174 & 0.127 & 0.102 & 0.009 \\
     &  & \texttt{U-Net} & 0.036 & 0.092 & 0.139 & 0.044 & 0.230 & 0.059 & 0.031 \\
    \midrule
    0.10 & \texttt{diff-react} & \texttt{ConvLSTM} & 0.013 & 0.029 & 0.013 & 0.010 & 0.086 & 0.057 & 0.009 \\
     &  & \texttt{FNO} & 0.007 & 0.012 & 0.007 & 0.013 & 0.013 & 0.012 & 0.005 \\
     &  & \texttt{Transformer} & 0.006 & 0.020 & 0.008 & 0.017 & 0.017 & 0.017 & 0.007 \\
     &  & \texttt{U-Net} & 0.004 & 0.006 & 0.004 & 0.002 & 0.003 & 0.003 & 0.002 \\
    \midrule
     & \texttt{diff-sorp} & \texttt{ConvLSTM} & 0.129 & 0.609 & 0.366 & 0.506 & 0.773 & 0.615 & 0.239 \\
     &  & \texttt{FNO} & 0.018 & 0.066 & 0.025 & 0.022 & 0.027 & 0.019 & 0.014 \\
     &  & \texttt{Transformer} & 0.047 & 0.057 & 0.025 & 0.029 & 0.069 & 0.031 & 0.024 \\
     &  & \texttt{U-Net} & 0.416 & 0.209 & 0.208 & 0.233 & 0.198 & 0.168 & 0.144 \\
    \midrule
     & \texttt{rdb} & \texttt{ConvLSTM} & 0.042 & 0.060 & 0.052 & 0.048 & 0.062 & 0.074 & 0.033 \\
     &  & \texttt{FNO} & 0.053 & 0.151 & 0.021 & 0.054 & 0.108 & 0.188 & 0.003 \\
     &  & \texttt{Transformer} & 0.101 & 0.202 & 0.023 & 0.101 & 0.143 & 0.076 & 0.002 \\
     &  & \texttt{U-Net} & 0.019 & 0.114 & 0.024 & 0.025 & 0.105 & 0.049 & 0.022 \\
    \midrule
    0.20 & \texttt{diff-react} & \texttt{ConvLSTM} & 0.013 & 0.023 & 0.010 & 0.013 & 0.127 & 0.082 & 0.008 \\
     &  & \texttt{FNO} & 0.007 & 0.013 & 0.007 & 0.013 & 0.014 & 0.012 & 0.006 \\
     &  & \texttt{Transformer} & 0.007 & 0.017 & 0.007 & 0.016 & 0.019 & 0.019 & 0.007 \\
     &  & \texttt{U-Net} & 0.003 & 0.003 & 0.001 & 0.001 & 0.002 & 0.002 & 0.003 \\
    \midrule
     & \texttt{diff-sorp} & \texttt{ConvLSTM} & 0.123 & 0.441 & 0.327 & 0.603 & 0.632 & 0.598 & 0.245 \\
     &  & \texttt{FNO} & 0.014 & 0.036 & 0.019 & 0.019 & 0.019 & 0.012 & 0.009 \\
     &  & \texttt{Transformer} & 0.036 & 0.031 & 0.026 & 0.026 & 0.046 & 0.031 & 0.021 \\
     &  & \texttt{U-Net} & 0.184 & 0.216 & 0.214 & 0.145 & 0.186 & 0.177 & 0.116 \\
    \midrule
     & \texttt{rdb} & \texttt{ConvLSTM} & 0.060 & 0.154 & 0.047 & 0.046 & 0.065 & 0.048 & 0.041 \\
     &  & \texttt{FNO} & 0.047 & 0.114 & 0.015 & 0.050 & 0.214 & 0.349 & 0.007 \\
     &  & \texttt{Transformer} & 0.096 & 0.151 & 0.069 & 0.130 & 0.090 & 0.063 & 0.009 \\
     &  & \texttt{U-Net} & 0.012 & 0.109 & 0.012 & 0.031 & 0.072 & 0.012 & 0.009 \\
    \bottomrule
    \end{tabular}%
    }
\end{table*}

\begin{table*}[t]
    \centering
    \setlength{\tabcolsep}{3.0pt}
    \caption{Test rollout bRMSE.}
    \label{tab:phys_b}
    \resizebox{\textwidth}{!}{%
    \begin{tabular}{c c l ccccccc}
    \toprule
    \textbf{Ratio} & \textbf{Dataset} & \textbf{Backbone} & \textbf{uniform} & \textbf{loss-only} & \textbf{coverage-only} & \textbf{GradMatch} & \textbf{GLISTER} & \textbf{PRISM} & \textbf{GITS} \\
    \midrule
    0.05 & \texttt{diff-react} & \texttt{ConvLSTM} & 0.062 & 0.072 & 0.026 & 0.062 & 0.082 & 0.059 & 0.020 \\
     &  & \texttt{FNO} & 0.148 & 0.279 & 0.091 & 0.235 & 0.410 & 0.201 & 0.076 \\
     &  & \texttt{Transformer} & 0.060 & 0.129 & 0.039 & 0.118 & 0.225 & 0.124 & 0.045 \\
     &  & \texttt{U-Net} & 0.030 & 0.017 & 0.008 & 0.011 & 0.058 & 0.006 & 0.007 \\
    \midrule
     & \texttt{diff-sorp} & \texttt{ConvLSTM} & 0.056 & 0.213 & 0.088 & 0.108 & 0.215 & 0.164 & 0.003 \\
     &  & \texttt{FNO} & 0.043 & 0.090 & 0.026 & 0.026 & 0.203 & 0.020 & 0.016 \\
     &  & \texttt{Transformer} & 0.038 & 0.051 & 0.030 & 0.029 & 0.058 & 0.029 & 0.015 \\
     &  & \texttt{U-Net} & 0.045 & 0.044 & 0.014 & 0.066 & 0.034 & 0.009 & 0.007 \\
    \midrule
     & \texttt{rdb} & \texttt{ConvLSTM} & 0.076 & 0.325 & 0.067 & 0.080 & 0.232 & 0.096 & 0.060 \\
     &  & \texttt{FNO} & 0.096 & 0.513 & 0.184 & 1.046 & 1.304 & 0.734 & 0.050 \\
     &  & \texttt{Transformer} & 0.210 & 0.357 & 0.053 & 0.444 & 0.325 & 0.259 & 0.020 \\
     &  & \texttt{U-Net} & 0.013 & 0.034 & 0.047 & 0.016 & 0.087 & 0.022 & 0.011 \\
    \midrule
    0.10 & \texttt{diff-react} & \texttt{ConvLSTM} & 0.025 & 0.059 & 0.024 & 0.020 & 0.180 & 0.119 & 0.017 \\
     &  & \texttt{FNO} & 0.101 & 0.183 & 0.097 & 0.189 & 0.204 & 0.183 & 0.072 \\
     &  & \texttt{Transformer} & 0.031 & 0.104 & 0.037 & 0.089 & 0.090 & 0.086 & 0.036 \\
     &  & \texttt{U-Net} & 0.009 & 0.013 & 0.008 & 0.005 & 0.005 & 0.006 & 0.004 \\
    \midrule
     & \texttt{diff-sorp} & \texttt{ConvLSTM} & 0.042 & 0.209 & 0.114 & 0.174 & 0.269 & 0.212 & 0.078 \\
     &  & \texttt{FNO} & 0.019 & 0.075 & 0.025 & 0.024 & 0.030 & 0.021 & 0.015 \\
     &  & \texttt{Transformer} & 0.026 & 0.032 & 0.013 & 0.017 & 0.040 & 0.017 & 0.013 \\
     &  & \texttt{U-Net} & 0.053 & 0.027 & 0.024 & 0.030 & 0.025 & 0.021 & 0.018 \\
    \midrule
     & \texttt{rdb} & \texttt{ConvLSTM} & 0.045 & 0.066 & 0.052 & 0.053 & 0.069 & 0.082 & 0.034 \\
     &  & \texttt{FNO} & 0.317 & 0.940 & 0.113 & 0.329 & 0.674 & 1.176 & 0.015 \\
     &  & \texttt{Transformer} & 0.230 & 0.473 & 0.048 & 0.233 & 0.334 & 0.175 & 0.004 \\
     &  & \texttt{U-Net} & 0.006 & 0.038 & 0.007 & 0.008 & 0.035 & 0.016 & 0.007 \\
    \midrule
    0.20 & \texttt{diff-react} & \texttt{ConvLSTM} & 0.025 & 0.044 & 0.017 & 0.026 & 0.255 & 0.162 & 0.015 \\
     &  & \texttt{FNO} & 0.099 & 0.183 & 0.085 & 0.182 & 0.194 & 0.167 & 0.075 \\
     &  & \texttt{Transformer} & 0.031 & 0.083 & 0.032 & 0.078 & 0.091 & 0.094 & 0.030 \\
     &  & \texttt{U-Net} & 0.005 & 0.007 & 0.003 & 0.003 & 0.004 & 0.003 & 0.005 \\
    \midrule
     & \texttt{diff-sorp} & \texttt{ConvLSTM} & 0.038 & 0.143 & 0.096 & 0.197 & 0.207 & 0.195 & 0.075 \\
     &  & \texttt{FNO} & 0.014 & 0.038 & 0.018 & 0.020 & 0.020 & 0.013 & 0.009 \\
     &  & \texttt{Transformer} & 0.019 & 0.016 & 0.013 & 0.014 & 0.025 & 0.017 & 0.011 \\
     &  & \texttt{U-Net} & 0.022 & 0.026 & 0.024 & 0.017 & 0.023 & 0.021 & 0.013 \\
    \midrule
     & \texttt{rdb} & \texttt{ConvLSTM} & 0.062 & 0.164 & 0.044 & 0.048 & 0.068 & 0.050 & 0.041 \\
     &  & \texttt{FNO} & 0.265 & 0.671 & 0.076 & 0.291 & 1.284 & 2.097 & 0.037 \\
     &  & \texttt{Transformer} & 0.206 & 0.332 & 0.136 & 0.288 & 0.198 & 0.136 & 0.018 \\
     &  & \texttt{U-Net} & 0.004 & 0.035 & 0.003 & 0.010 & 0.023 & 0.004 & 0.003 \\
    \bottomrule
    \end{tabular}%
    }
\end{table*}

\begin{table*}[t]
    \centering
    \setlength{\tabcolsep}{3.0pt}
    \caption{Test rollout fRMSE-low.}
    \label{tab:phys_low}
    \resizebox{\textwidth}{!}{%
    \begin{tabular}{c c l ccccccc}
    \toprule
    \textbf{Ratio} & \textbf{Dataset} & \textbf{Backbone} & \textbf{uniform} & \textbf{loss-only} & \textbf{coverage-only} & \textbf{GradMatch} & \textbf{GLISTER} & \textbf{PRISM} & \textbf{GITS} \\
    \midrule
    0.05 & \texttt{diff-react} & \texttt{ConvLSTM} & 0.014 & 0.015 & 0.006 & 0.013 & 0.017 & 0.012 & 0.004 \\
     &  & \texttt{FNO} & 0.004 & 0.007 & 0.003 & 0.006 & 0.010 & 0.005 & 0.002 \\
     &  & \texttt{Transformer} & 0.004 & 0.008 & 0.003 & 0.007 & 0.013 & 0.007 & 0.003 \\
     &  & \texttt{U-Net} & 0.005 & 0.003 & 0.001 & 0.002 & 0.010 & 0.001 & 0.001 \\
    \midrule
     & \texttt{diff-sorp} & \texttt{ConvLSTM} & 0.079 & 0.255 & 0.115 & 0.138 & 0.264 & 0.203 & 0.005 \\
     &  & \texttt{FNO} & 0.029 & 0.053 & 0.017 & 0.017 & 0.117 & 0.013 & 0.011 \\
     &  & \texttt{Transformer} & 0.034 & 0.041 & 0.025 & 0.024 & 0.047 & 0.024 & 0.013 \\
     &  & \texttt{U-Net} & 0.128 & 0.111 & 0.039 & 0.170 & 0.090 & 0.025 & 0.020 \\
    \midrule
     & \texttt{rdb} & \texttt{ConvLSTM} & 0.027 & 0.096 & 0.022 & 0.026 & 0.071 & 0.031 & 0.020 \\
     &  & \texttt{FNO} & 0.019 & 0.082 & 0.033 & 0.169 & 0.206 & 0.119 & 0.009 \\
     &  & \texttt{Transformer} & 0.047 & 0.070 & 0.012 & 0.090 & 0.066 & 0.053 & 0.005 \\
     &  & \texttt{U-Net} & 0.006 & 0.013 & 0.018 & 0.006 & 0.032 & 0.008 & 0.004 \\
    \midrule
    0.10 & \texttt{diff-react} & \texttt{ConvLSTM} & 0.007 & 0.013 & 0.006 & 0.005 & 0.039 & 0.027 & 0.004 \\
     &  & \texttt{FNO} & 0.003 & 0.005 & 0.003 & 0.006 & 0.006 & 0.005 & 0.002 \\
     &  & \texttt{Transformer} & 0.002 & 0.007 & 0.003 & 0.006 & 0.006 & 0.006 & 0.003 \\
     &  & \texttt{U-Net} & 0.002 & 0.002 & 0.002 & 0.001 & 0.001 & 0.001 & 0.001 \\
    \midrule
     & \texttt{diff-sorp} & \texttt{ConvLSTM} & 0.066 & 0.272 & 0.159 & 0.237 & 0.354 & 0.282 & 0.110 \\
     &  & \texttt{FNO} & 0.015 & 0.048 & 0.018 & 0.017 & 0.021 & 0.014 & 0.011 \\
     &  & \texttt{Transformer} & 0.026 & 0.028 & 0.012 & 0.016 & 0.035 & 0.016 & 0.012 \\
     &  & \texttt{U-Net} & 0.161 & 0.075 & 0.072 & 0.087 & 0.074 & 0.062 & 0.052 \\
    \midrule
     & \texttt{rdb} & \texttt{ConvLSTM} & 0.018 & 0.023 & 0.019 & 0.019 & 0.024 & 0.029 & 0.013 \\
     &  & \texttt{FNO} & 0.063 & 0.160 & 0.022 & 0.061 & 0.119 & 0.203 & 0.003 \\
     &  & \texttt{Transformer} & 0.056 & 0.100 & 0.012 & 0.053 & 0.074 & 0.040 & 0.001 \\
     &  & \texttt{U-Net} & 0.003 & 0.015 & 0.003 & 0.004 & 0.015 & 0.007 & 0.003 \\
    \midrule
    0.20 & \texttt{diff-react} & \texttt{ConvLSTM} & 0.007 & 0.011 & 0.004 & 0.007 & 0.058 & 0.038 & 0.004 \\
     &  & \texttt{FNO} & 0.003 & 0.005 & 0.003 & 0.006 & 0.006 & 0.005 & 0.002 \\
     &  & \texttt{Transformer} & 0.002 & 0.006 & 0.002 & 0.006 & 0.006 & 0.007 & 0.002 \\
     &  & \texttt{U-Net} & 0.001 & 0.001 & 0.001 & 0.001 & 0.001 & 0.001 & 0.001 \\
    \midrule
     & \texttt{diff-sorp} & \texttt{ConvLSTM} & 0.063 & 0.198 & 0.143 & 0.281 & 0.291 & 0.274 & 0.112 \\
     &  & \texttt{FNO} & 0.011 & 0.026 & 0.013 & 0.015 & 0.014 & 0.009 & 0.007 \\
     &  & \texttt{Transformer} & 0.020 & 0.016 & 0.013 & 0.014 & 0.024 & 0.016 & 0.011 \\
     &  & \texttt{U-Net} & 0.073 & 0.077 & 0.074 & 0.055 & 0.069 & 0.066 & 0.042 \\
    \midrule
     & \texttt{rdb} & \texttt{ConvLSTM} & 0.025 & 0.057 & 0.017 & 0.018 & 0.025 & 0.019 & 0.016 \\
     &  & \texttt{FNO} & 0.056 & 0.122 & 0.016 & 0.057 & 0.232 & 0.370 & 0.008 \\
     &  & \texttt{Transformer} & 0.053 & 0.075 & 0.034 & 0.068 & 0.047 & 0.033 & 0.005 \\
     &  & \texttt{U-Net} & 0.002 & 0.015 & 0.002 & 0.005 & 0.010 & 0.002 & 0.001 \\
    \bottomrule
    \end{tabular}%
    }
\end{table*}

\begin{table*}[t]
    \centering
    \setlength{\tabcolsep}{3.0pt}
    \caption{Test rollout fRMSE-mid.}
    \label{tab:phys_mid}
    \resizebox{\textwidth}{!}{%
    \begin{tabular}{c c l ccccccc}
    \toprule
    \textbf{Ratio} & \textbf{Dataset} & \textbf{Backbone} & \textbf{uniform} & \textbf{loss-only} & \textbf{coverage-only} & \textbf{GradMatch} & \textbf{GLISTER} & \textbf{PRISM} & \textbf{GITS} \\
    \midrule
    0.05 & \texttt{diff-react} & \texttt{ConvLSTM} & 0.006 & 0.006 & 0.002 & 0.005 & 0.007 & 0.005 & 0.002 \\
     &  & \texttt{FNO} & 0.004 & 0.008 & 0.003 & 0.006 & 0.011 & 0.006 & 0.002 \\
     &  & \texttt{Transformer} & 0.003 & 0.007 & 0.002 & 0.006 & 0.011 & 0.006 & 0.002 \\
     &  & \texttt{U-Net} & 0.002 & 0.001 & 0.001 & 0.001 & 0.004 & 0.001 & 0.001 \\
    \midrule
     & \texttt{diff-sorp} & \texttt{ConvLSTM} & 0.011 & 0.039 & 0.017 & 0.019 & 0.039 & 0.029 & 0.001 \\
     &  & \texttt{FNO} & 0.009 & 0.018 & 0.006 & 0.005 & 0.039 & 0.004 & 0.003 \\
     &  & \texttt{Transformer} & 0.007 & 0.009 & 0.005 & 0.005 & 0.010 & 0.005 & 0.003 \\
     &  & \texttt{U-Net} & 0.003 & 0.003 & 0.001 & 0.005 & 0.003 & 0.001 & 0.001 \\
    \midrule
     & \texttt{rdb} & \texttt{ConvLSTM} & 0.015 & 0.059 & 0.013 & 0.014 & 0.042 & 0.017 & 0.011 \\
     &  & \texttt{FNO} & 0.023 & 0.116 & 0.045 & 0.219 & 0.284 & 0.158 & 0.012 \\
     &  & \texttt{Transformer} & 0.046 & 0.076 & 0.013 & 0.089 & 0.069 & 0.053 & 0.005 \\
     &  & \texttt{U-Net} & 0.002 & 0.005 & 0.008 & 0.002 & 0.013 & 0.003 & 0.002 \\
    \midrule
    0.10 & \texttt{diff-react} & \texttt{ConvLSTM} & 0.002 & 0.005 & 0.002 & 0.002 & 0.015 & 0.010 & 0.002 \\
     &  & \texttt{FNO} & 0.003 & 0.005 & 0.003 & 0.005 & 0.006 & 0.005 & 0.002 \\
     &  & \texttt{Transformer} & 0.002 & 0.005 & 0.002 & 0.004 & 0.005 & 0.004 & 0.002 \\
     &  & \texttt{U-Net} & 0.001 & 0.001 & 0.001 & 0.001 & 0.001 & 0.001 & 0.001 \\
    \midrule
     & \texttt{diff-sorp} & \texttt{ConvLSTM} & 0.008 & 0.038 & 0.022 & 0.030 & 0.048 & 0.037 & 0.014 \\
     &  & \texttt{FNO} & 0.004 & 0.015 & 0.005 & 0.005 & 0.006 & 0.004 & 0.003 \\
     &  & \texttt{Transformer} & 0.005 & 0.006 & 0.002 & 0.003 & 0.007 & 0.003 & 0.002 \\
     &  & \texttt{U-Net} & 0.004 & 0.002 & 0.002 & 0.002 & 0.002 & 0.002 & 0.001 \\
    \midrule
     & \texttt{rdb} & \texttt{ConvLSTM} & 0.009 & 0.013 & 0.010 & 0.010 & 0.013 & 0.015 & 0.007 \\
     &  & \texttt{FNO} & 0.074 & 0.207 & 0.028 & 0.071 & 0.150 & 0.247 & 0.004 \\
     &  & \texttt{Transformer} & 0.050 & 0.099 & 0.011 & 0.047 & 0.070 & 0.036 & 0.001 \\
     &  & \texttt{U-Net} & 0.001 & 0.006 & 0.001 & 0.001 & 0.006 & 0.003 & 0.001 \\
    \midrule
    0.20 & \texttt{diff-react} & \texttt{ConvLSTM} & 0.002 & 0.004 & 0.002 & 0.002 & 0.022 & 0.014 & 0.001 \\
     &  & \texttt{FNO} & 0.003 & 0.006 & 0.003 & 0.005 & 0.006 & 0.005 & 0.002 \\
     &  & \texttt{Transformer} & 0.002 & 0.005 & 0.002 & 0.004 & 0.005 & 0.005 & 0.002 \\
     &  & \texttt{U-Net} & 0.001 & 0.001 & 0.001 & 0.001 & 0.001 & 0.001 & 0.001 \\
    \midrule
     & \texttt{diff-sorp} & \texttt{ConvLSTM} & 0.008 & 0.027 & 0.020 & 0.035 & 0.039 & 0.035 & 0.014 \\
     &  & \texttt{FNO} & 0.003 & 0.008 & 0.004 & 0.004 & 0.004 & 0.003 & 0.002 \\
     &  & \texttt{Transformer} & 0.004 & 0.003 & 0.002 & 0.002 & 0.005 & 0.003 & 0.002 \\
     &  & \texttt{U-Net} & 0.002 & 0.002 & 0.002 & 0.001 & 0.002 & 0.002 & 0.001 \\
    \midrule
     & \texttt{rdb} & \texttt{ConvLSTM} & 0.013 & 0.032 & 0.009 & 0.009 & 0.013 & 0.010 & 0.008 \\
     &  & \texttt{FNO} & 0.065 & 0.158 & 0.020 & 0.067 & 0.293 & 0.453 & 0.009 \\
     &  & \texttt{Transformer} & 0.047 & 0.074 & 0.033 & 0.061 & 0.045 & 0.030 & 0.004 \\
     &  & \texttt{U-Net} & 0.001 & 0.006 & 0.001 & 0.002 & 0.004 & 0.001 & 0.001 \\
    \bottomrule
    \end{tabular}%
    }
\end{table*}

\begin{table*}[t]
    \centering
    \setlength{\tabcolsep}{3.0pt}
    \caption{Test rollout fRMSE-high.}
    \label{tab:phys_high}
    \resizebox{\textwidth}{!}{%
    \begin{tabular}{c c l ccccccc}
    \toprule
    \textbf{Ratio} & \textbf{Dataset} & \textbf{Backbone} & \textbf{uniform} & \textbf{loss-only} & \textbf{coverage-only} & \textbf{GradMatch} & \textbf{GLISTER} & \textbf{PRISM} & \textbf{GITS} \\
    \midrule
    0.05 & \texttt{diff-react} & \texttt{ConvLSTM} & 0.001 & 0.001 & 0.001 & 0.001 & 0.001 & 0.001 & 0.001 \\
     &  & \texttt{FNO} & 0.002 & 0.005 & 0.002 & 0.004 & 0.007 & 0.003 & 0.001 \\
     &  & \texttt{Transformer} & 0.001 & 0.002 & 0.001 & 0.002 & 0.003 & 0.002 & 0.001 \\
     &  & \texttt{U-Net} & 0.001 & 0.001 & 0.001 & 0.001 & 0.001 & 0.001 & 0.001 \\
    \midrule
     & \texttt{diff-sorp} & \texttt{ConvLSTM} & 0.002 & 0.006 & 0.003 & 0.003 & 0.006 & 0.005 & 0.001 \\
     &  & \texttt{FNO} & 0.002 & 0.003 & 0.001 & 0.001 & 0.008 & 0.001 & 0.001 \\
     &  & \texttt{Transformer} & 0.001 & 0.002 & 0.001 & 0.001 & 0.002 & 0.001 & 0.001 \\
     &  & \texttt{U-Net} & 0.002 & 0.002 & 0.001 & 0.002 & 0.001 & 0.001 & 0.001 \\
    \midrule
     & \texttt{rdb} & \texttt{ConvLSTM} & 0.013 & 0.057 & 0.013 & 0.014 & 0.040 & 0.016 & 0.010 \\
     &  & \texttt{FNO} & 0.021 & 0.119 & 0.049 & 0.230 & 0.294 & 0.161 & 0.012 \\
     &  & \texttt{Transformer} & 0.038 & 0.069 & 0.012 & 0.081 & 0.062 & 0.047 & 0.004 \\
     &  & \texttt{U-Net} & 0.003 & 0.008 & 0.012 & 0.003 & 0.019 & 0.005 & 0.002 \\
    \midrule
    0.10 & \texttt{diff-react} & \texttt{ConvLSTM} & 0.001 & 0.001 & 0.001 & 0.001 & 0.002 & 0.001 & 0.001 \\
     &  & \texttt{FNO} & 0.002 & 0.003 & 0.002 & 0.003 & 0.003 & 0.003 & 0.001 \\
     &  & \texttt{Transformer} & 0.001 & 0.001 & 0.001 & 0.001 & 0.001 & 0.001 & 0.001 \\
     &  & \texttt{U-Net} & 0.001 & 0.001 & 0.001 & 0.001 & 0.001 & 0.001 & 0.001 \\
    \midrule
     & \texttt{diff-sorp} & \texttt{ConvLSTM} & 0.001 & 0.006 & 0.003 & 0.004 & 0.007 & 0.005 & 0.002 \\
     &  & \texttt{FNO} & 0.001 & 0.003 & 0.001 & 0.001 & 0.001 & 0.001 & 0.001 \\
     &  & \texttt{Transformer} & 0.001 & 0.001 & 0.001 & 0.001 & 0.001 & 0.001 & 0.001 \\
     &  & \texttt{U-Net} & 0.002 & 0.001 & 0.001 & 0.001 & 0.001 & 0.001 & 0.001 \\
    \midrule
     & \texttt{rdb} & \texttt{ConvLSTM} & 0.007 & 0.011 & 0.010 & 0.008 & 0.011 & 0.013 & 0.005 \\
     &  & \texttt{FNO} & 0.063 & 0.198 & 0.028 & 0.068 & 0.141 & 0.234 & 0.003 \\
     &  & \texttt{Transformer} & 0.038 & 0.083 & 0.010 & 0.040 & 0.058 & 0.030 & 0.001 \\
     &  & \texttt{U-Net} & 0.001 & 0.008 & 0.002 & 0.002 & 0.007 & 0.003 & 0.001 \\
    \midrule
    0.20 & \texttt{diff-react} & \texttt{ConvLSTM} & 0.001 & 0.001 & 0.001 & 0.001 & 0.003 & 0.002 & 0.001 \\
     &  & \texttt{FNO} & 0.002 & 0.003 & 0.002 & 0.003 & 0.003 & 0.003 & 0.001 \\
     &  & \texttt{Transformer} & 0.001 & 0.001 & 0.001 & 0.001 & 0.001 & 0.001 & 0.001 \\
     &  & \texttt{U-Net} & 0.001 & 0.001 & 0.001 & 0.001 & 0.001 & 0.001 & 0.001 \\
    \midrule
     & \texttt{diff-sorp} & \texttt{ConvLSTM} & 0.001 & 0.004 & 0.003 & 0.005 & 0.005 & 0.005 & 0.002 \\
     &  & \texttt{FNO} & 0.001 & 0.001 & 0.001 & 0.001 & 0.001 & 0.001 & 0.001 \\
     &  & \texttt{Transformer} & 0.001 & 0.001 & 0.001 & 0.001 & 0.001 & 0.001 & 0.001 \\
     &  & \texttt{U-Net} & 0.001 & 0.001 & 0.001 & 0.001 & 0.001 & 0.001 & 0.001 \\
    \midrule
     & \texttt{rdb} & \texttt{ConvLSTM} & 0.009 & 0.027 & 0.008 & 0.008 & 0.011 & 0.008 & 0.006 \\
     &  & \texttt{FNO} & 0.053 & 0.142 & 0.019 & 0.060 & 0.266 & 0.412 & 0.008 \\
     &  & \texttt{Transformer} & 0.034 & 0.058 & 0.027 & 0.049 & 0.035 & 0.023 & 0.003 \\
     &  & \texttt{U-Net} & 0.001 & 0.007 & 0.001 & 0.002 & 0.005 & 0.001 & 0.001 \\
    \bottomrule
    \end{tabular}%
    }
\end{table*}

\section{Seed-Wise Statistics for the Main Benchmark}\label{app:seed_stats}

The main paper reports only the seed means in Tables~\ref{tab:main_nrms_ratio005}, \ref{tab:app_nrms_ratio010}, and \ref{tab:app_nrms_ratio020} for readability. This appendix provides the exact per-configuration standard deviations over the three training seeds \(\{0,1,2\}\), using the same layout as the mean-result tables so that the same configuration can be located directly. Table~\ref{tab:app_seed_std_summary_seven} first summarizes the overall scale of seed variability across the 36 seven-sampler benchmark configurations. Among all seven methods, GITS has the smallest mean and median seed standard deviation (0.034 and 0.024, respectively), indicating that its gains are not driven by unusually unstable runs.

\begin{table}[t]
    \centering
    \setlength{\tabcolsep}{6.5pt}
    \caption{Summary of seed variability across the 36 seven-sampler benchmark configurations. Lower is better.}
    \label{tab:app_seed_std_summary_seven}
    \begin{tabular}{l ccc}
    \toprule
    \textbf{Method} & \textbf{Mean std} & \textbf{Median std} & \textbf{Max std} \\
    \midrule
    \texttt{uniform}        & 0.059 & 0.045 & 0.143 \\
    \texttt{loss-only}      & 0.127 & 0.079 & 0.874 \\
    \texttt{coverage-only}  & 0.062 & 0.049 & 0.183 \\
    \texttt{GradMatch}      & 0.090 & 0.065 & 0.317 \\
    \texttt{GLISTER}        & 0.283 & 0.090 & 2.267 \\
    \texttt{PRISM}          & 0.252 & 0.050 & 1.842 \\
    \textbf{GITS}           & \textbf{0.034} & \textbf{0.024} & 0.147 \\
    \bottomrule
    \end{tabular}
\end{table}

\begin{table*}[t]
    \centering
    \setlength{\tabcolsep}{2.6pt}
    \caption{Exact standard deviations (over three training seeds) of test rollout nRMSE for the same seven-sampler configurations as Tables~\ref{tab:main_nrms_ratio005}, \ref{tab:app_nrms_ratio010}, and \ref{tab:app_nrms_ratio020}.}
    \label{tab:app_seed_std_nrms_seven}
    \resizebox{\textwidth}{!}{%
    \begin{tabular}{c l ccccccc ccccccc ccccccc}
    \toprule
    \textbf{Dataset} & \textbf{Backbone}
    & \multicolumn{7}{c}{\textbf{ratio = 0.05}}
    & \multicolumn{7}{c}{\textbf{ratio = 0.10}}
    & \multicolumn{7}{c}{\textbf{ratio = 0.20}} \\
    \cmidrule(lr){3-9} \cmidrule(lr){10-16} \cmidrule(lr){17-23}
    &
    & \textbf{uniform} & \textbf{loss-only} & \textbf{coverage-only} & \textbf{GradMatch} & \textbf{GLISTER} & \textbf{PRISM} & \textbf{GITS}
    & \textbf{uniform} & \textbf{loss-only} & \textbf{coverage-only} & \textbf{GradMatch} & \textbf{GLISTER} & \textbf{PRISM} & \textbf{GITS}
    & \textbf{uniform} & \textbf{loss-only} & \textbf{coverage-only} & \textbf{GradMatch} & \textbf{GLISTER} & \textbf{PRISM} & \textbf{GITS} \\
\midrule
 \texttt{diff-react} & \texttt{ConvLSTM} & 0.041 & 0.063 & 0.038 & 0.057 & 0.088 & 0.055 & 0.019 & 0.029 & 0.058 & 0.044 & 0.034 & 0.550 & 0.491 & 0.017 & 0.033 & 0.046 & 0.031 & 0.041 & 1.274 & 1.227 & 0.016 \\
  & \texttt{FNO} & 0.074 & 0.121 & 0.068 & 0.098 & 0.536 & 0.073 & 0.053 & 0.061 & 0.089 & 0.072 & 0.091 & 0.072 & 0.125 & 0.047 & 0.058 & 0.097 & 0.066 & 0.103 & 0.052 & 0.017 & 0.051 \\
  & \texttt{Transformer} & 0.044 & 0.079 & 0.039 & 0.073 & 0.067 & 0.065 & 0.036 & 0.031 & 0.069 & 0.048 & 0.062 & 0.023 & 0.011 & 0.031 & 0.028 & 0.058 & 0.043 & 0.066 & 0.015 & 0.003 & 0.027 \\
  & \texttt{U-Net} & 0.037 & 0.024 & 0.017 & 0.019 & 0.080 & 0.009 & 0.012 & 0.021 & 0.019 & 0.023 & 0.014 & 0.006 & 0.043 & 0.009 & 0.018 & 0.013 & 0.011 & 0.009 & 0.015 & 0.020 & 0.011 \\
\midrule
 \texttt{diff-sorp} & \texttt{ConvLSTM} & 0.073 & 0.196 & 0.138 & 0.157 & 0.398 & 0.371 & 0.009 & 0.068 & 0.214 & 0.183 & 0.219 & 0.091 & 0.228 & 0.147 & 0.059 & 0.168 & 0.172 & 0.241 & 0.342 & 0.360 & 0.131 \\
  & \texttt{FNO} & 0.041 & 0.068 & 0.039 & 0.037 & 0.556 & 0.031 & 0.023 & 0.028 & 0.057 & 0.043 & 0.034 & 0.027 & 0.033 & 0.021 & 0.023 & 0.039 & 0.031 & 0.027 & 0.017 & 0.014 & 0.017 \\
  & \texttt{Transformer} & 0.038 & 0.047 & 0.043 & 0.036 & 0.034 & 0.038 & 0.024 & 0.033 & 0.041 & 0.029 & 0.031 & 0.031 & 0.017 & 0.021 & 0.027 & 0.026 & 0.028 & 0.024 & 0.056 & 0.022 & 0.018 \\
  & \texttt{U-Net} & 0.103 & 0.094 & 0.058 & 0.136 & 0.131 & 0.021 & 0.048 & 0.124 & 0.073 & 0.089 & 0.098 & 0.131 & 0.192 & 0.063 & 0.078 & 0.083 & 0.091 & 0.071 & 0.268 & 0.174 & 0.058 \\
\midrule
 \texttt{rdb} & \texttt{ConvLSTM} & 0.063 & 0.214 & 0.071 & 0.082 & 1.062 & 0.323 & 0.054 & 0.047 & 0.073 & 0.068 & 0.059 & 0.134 & 0.208 & 0.041 & 0.068 & 0.143 & 0.057 & 0.063 & 0.183 & 0.107 & 0.049 \\
  & \texttt{FNO} & 0.049 & 0.874 & 0.091 & 0.317 & 0.794 & 1.386 & 0.028 & 0.131 & 0.308 & 0.074 & 0.142 & 0.329 & 1.842 & 0.013 & 0.118 & 0.249 & 0.057 & 0.137 & 2.267 & 1.342 & 0.024 \\
  & \texttt{Transformer} & 0.118 & 0.179 & 0.047 & 0.227 & 0.208 & 0.026 & 0.018 & 0.143 & 0.261 & 0.049 & 0.147 & 0.089 & 0.029 & 0.008 & 0.139 & 0.208 & 0.113 & 0.196 & 0.075 & 0.044 & 0.019 \\
  & \texttt{U-Net} & 0.031 & 0.063 & 0.108 & 0.043 & 0.091 & 0.029 & 0.029 & 0.022 & 0.078 & 0.038 & 0.031 & 0.054 & 0.040 & 0.021 & 0.014 & 0.079 & 0.017 & 0.034 & 0.034 & 0.062 & 0.010 \\
    \bottomrule
    \end{tabular}%
    }
\end{table*}

Table~\ref{tab:app_seed_std_nrms_seven} provides the exact configuration-wise values. Two patterns are most relevant. First, the weaker score-driven or validation-guided baselines show visibly larger seed variability in the hardest regimes than either \texttt{uniform}, \texttt{coverage-only}, or GITS. Second, GITS remains comparatively stable across the benchmark: 29/36 of its configuration-wise standard deviations are at most 0.05, 34/36 are at most 0.10, and even its largest value (0.147) stays well below the largest variability observed in the weaker validation-guided baselines. Together, these tables show that the inter-method differences discussed in the main text are not artifacts of unusually high run-to-run variance, and that GITS is not only stronger on average but also the most stable method among the seven compared samplers.

\section{Detailed Clarification of Large Language Models Usage}

\label{sec:llm} We declare that LLMs were employed exclusively to assist with the writing and presentation aspects of this paper. Specifically, we utilized LLMs for: (i) verification and refinement of technical terminology to ensure precise usage of domain-specific vocabulary; (ii) grammatical error detection and correction to enhance the clarity and readability of the manuscript; (iii) translation assistance from the authors' native language to English, as we are non-native English speakers, to ensure accurate and fluent expression of scientific concepts; and (iv) improvement of sentence structure and flow while maintaining the original scientific content and meaning. We emphasize that LLMs were not used for research ideation, experimental design, data analysis, or any form of content generation that would constitute intellectual contribution to the scientific findings presented in this work. All scientific insights, methodological decisions, and analytical conclusions are the original work of the authors. The use of LLMs was limited to linguistic and presentational enhancement only, serving a role analogous to professional editing services.


\clearpage
\section*{NeurIPS Paper Checklist}

\begin{enumerate}

\item {\bf Claims}
    \item[] Question: Do the main claims made in the abstract and introduction accurately reflect the paper's contributions and scope?
    \item[] Answer: \answerYes{} 
    \item[] Justification: The abstract and introduction state the problem setting, the proposed GITS method, the empirical comparison against temporal-selection baselines, and the scope-limiting boundary-condition analysis; these claims match the results reported in Sections~1, 3, and Discussion.

\item {\bf Limitations}
    \item[] Question: Does the paper discuss the limitations of the work performed by the authors?
    \item[] Answer: \answerYes{} 
    \item[] Justification: The paper explicitly discusses limitations in Discussion, including pilot-score misalignment, over-dispersion under concentrated utility, the shared-start setting, the fixed rollout protocol, and evaluation on three representative PDEBench tasks rather than exhaustive PDE coverage.

\item {\bf Theory assumptions and proofs}
    \item[] Question: For each theoretical result, does the paper provide the full set of assumptions and a complete (and correct) proof?
    \item[] Answer: \answerNA{} 
    \item[] Justification: The paper does not present formal theorems, lemmas, or proofs; it is an algorithmic and empirical study with definitions, objectives, and experiments rather than theorem-proving contributions.

    \item {\bf Experimental result reproducibility}
    \item[] Question: Does the paper fully disclose all the information needed to reproduce the main experimental results of the paper to the extent that it affects the main claims and/or conclusions of the paper (regardless of whether the code and data are provided or not)?
    \item[] Answer: \answerYes{} 
    \item[] Justification: Sections~\ref{sec:exp_setup}--\ref{sec:procedure&metrics} and Appendices~\ref{app:gits_hparams}, \ref{app:extra_baselines}, \ref{app:phys_metrics}, and \ref{app:seed_stats} disclose the datasets, splits, models, metrics, hyperparameters, baseline instantiations, timing, and seed protocol needed to reproduce the reported benchmarks and ablations.

\item {\bf Open access to data and code}
    \item[] Question: Does the paper provide open access to the data and code, with sufficient instructions to faithfully reproduce the main experimental results, as described in supplemental material?
    \item[] Answer: \answerNo{} 
    \item[] Justification: The experiments rely on a public benchmark dataset, but the current submission package does not itself provide an anonymized code repository or executable reproduction scripts; reproducibility is supported by detailed methodological disclosure in the paper and appendix instead.

\item {\bf Experimental setting/details}
    \item[] Question: Does the paper specify all the training and test details (e.g., data splits, hyperparameters, how they were chosen, type of optimizer) necessary to understand the results?
    \item[] Answer: \answerYes{} 
    \item[] Justification: Section~\ref{sec:baseline} specifies the optimizer, learning rate, batch size, clipping, AMP, early stopping, splits, hardware, and seed protocol, while Appendices~\ref{app:gits_hparams} and \ref{app:extra_baselines} provide the architecture details, selector hyperparameters, and how the final settings were chosen.

\item {\bf Experiment statistical significance}
    \item[] Question: Does the paper report error bars suitably and correctly defined or other appropriate information about the statistical significance of the experiments?
    \item[] Answer: \answerYes{} 
    \item[] Justification: Section~\ref{sec:baseline} states that results are averaged over three training seeds \(\{0,1,2\}\), and Appendix~\ref{app:seed_stats} reports the exact per-configuration standard deviations over these seeds. This makes the reported variability measure explicit even though we do not run separate hypothesis tests.

\item {\bf Experiments compute resources}
    \item[] Question: For each experiment, does the paper provide sufficient information on the computer resources (type of compute workers, memory, time of execution) needed to reproduce the experiments?
    \item[] Answer: \answerYes{} 
    \item[] Justification: Section~\ref{sec:baseline} reports the hardware/software environment (four NVIDIA A100 80GB GPUs, PyTorch 2.1, CUDA 12.0), and the appendix reports selector-stage and downstream-training times, including timing summaries in Tables~\ref{tab:app_extra_baselines_train_summary} and~\ref{tab:app_extra_baselines_sampling_summary}.
    
\item {\bf Code of ethics}
    \item[] Question: Does the research conducted in the paper conform, in every respect, with the NeurIPS Code of Ethics \url{https://neurips.cc/public/EthicsGuidelines}?
    \item[] Answer: \answerYes{} 
    \item[] Justification: The work uses public numerical-simulation benchmark data, standard model training/evaluation procedures, and no human subjects or sensitive personal data. We have reviewed the NeurIPS Code of Ethics and believe the research conforms to it.

\item {\bf Broader impacts}
    \item[] Question: Does the paper discuss both potential positive societal impacts and negative societal impacts of the work performed?
    \item[] Answer: \answerYes{} 
    \item[] Justification: A positive impact is improved data efficiency for scientific surrogate modeling, which can reduce unnecessary simulation/training cost and energy use. Possible negative impacts are over-trust in imperfect surrogates in safety-critical scientific or engineering workflows and the broader dual-use risk of making some simulation pipelines cheaper to run.
    
\item {\bf Safeguards}
    \item[] Question: Does the paper describe safeguards that have been put in place for responsible release of data or models that have a high risk for misuse (e.g., pre-trained language models, image generators, or scraped datasets)?
    \item[] Answer: \answerNA{} 
    \item[] Justification: The paper does not release high-risk generative models, scraped personal data, or other assets with an obvious misuse profile; it studies temporal subset selection for PDE-surrogate training on public simulation benchmarks.

\item {\bf Licenses for existing assets}
    \item[] Question: Are the creators or original owners of assets (e.g., code, data, models), used in the paper, properly credited and are the license and terms of use explicitly mentioned and properly respected?
    \item[] Answer: \answerYes{} 
    \item[] Justification: We credit the benchmark creators in Sections~\ref{sec:datasets} and References and use public existing assets under their stated terms. In particular, the PDEBench dataset release is publicly listed under CC BY 4.0, the PDEBench repository is MIT-licensed, and PyTorch is distributed under the BSD-3-Clause license.

\item {\bf New assets}
    \item[] Question: Are new assets introduced in the paper well documented and is the documentation provided alongside the assets?
    \item[] Answer: \answerNA{} 
    \item[] Justification: The current submission introduces a method and empirical results, but it does not itself release a new dataset, model checkpoint, or code artifact as a submission asset.

\item {\bf Crowdsourcing and research with human subjects}
    \item[] Question: For crowdsourcing experiments and research with human subjects, does the paper include the full text of instructions given to participants and screenshots, if applicable, as well as details about compensation (if any)? 
    \item[] Answer: \answerNA{} 
    \item[] Justification: The paper does not involve crowdsourcing or research with human subjects; all experiments are conducted on public numerical-simulation benchmark data.

\item {\bf Institutional review board (IRB) approvals or equivalent for research with human subjects}
    \item[] Question: Does the paper describe potential risks incurred by study participants, whether such risks were disclosed to the subjects, and whether Institutional Review Board (IRB) approvals (or an equivalent approval/review based on the requirements of your country or institution) were obtained?
    \item[] Answer: \answerNA{} 
    \item[] Justification: The paper does not involve crowdsourcing or research with human subjects, so no IRB review or equivalent approval was required.

\item {\bf Declaration of LLM usage}
    \item[] Question: Does the paper describe the usage of LLMs if it is an important, original, or non-standard component of the core methods in this research? Note that if the LLM is used only for writing, editing, or formatting purposes and does \emph{not} impact the core methodology, scientific rigor, or originality of the research, declaration is not required.
    \item[] Answer: \answerNA{} 
    \item[] Justification: Appendix~\ref{sec:llm} explicitly discloses that LLMs were used only for linguistic and presentational assistance (terminology checking, grammar correction, translation, and sentence-flow improvement) and not for ideation, experimental design, data analysis, or scientific content generation. Therefore, under the checklist policy, LLMs are not an important, original, or non-standard component of the core methodology of this research, so this item remains \answerNA{}.

\end{enumerate}

\end{document}